\documentclass[twoside,11pt]{article}

\usepackage[preprint]{jmlr2e}
\usepackage{fullpage}
\usepackage{enumerate}
\usepackage{color}

\usepackage{etoolbox}
\usepackage{microtype}    %
\usepackage{comment}
\usepackage{abstract}     %
\usepackage{amsmath,amssymb,mathtools,amsfonts}

\def\beq{\begin{equation} }\def\eeq{\end{equation} }\def\ep{\varepsilon}\def\1{\mathbf{1}}

\usepackage{amsmath,amssymb,mathtools,amsfonts}
\usepackage{subfigure}
\usepackage{enumerate}
\newcommand{\RR}{\mathbb{R}}
\newcommand{\EE}{\mathbb{E}}
\newcommand{\bg}{\bm{g}}
\newcommand{\bG}{\bm{G}}
\newcommand{\bX}{\bm{X}}
\newcommand{\bF}{\bm{F}}
\newcommand{\cO}{\mathcal{O}}
\newcommand{\cF}{\mathcal{F}}

\newtheorem{assumption}[theorem]{Assumption}
\usepackage{algorithm,algorithmic}
\newcommand{\cS}{\mathcal{S}} 
\RequirePackage{bm}

\begin{document}

\title{Stochastic Recursive Variance Reduction for Efficient Smooth Non-Convex Compositional Optimization}

\author{\name 
Huizhuo Yuan \email huizhuo.yuan@gmail.com
 \\
       \addr School of Mathematical Sciences\\
       Peking University\\
			 Beijing, CHN
\AND
       \name 
Xiangru Lian \email admin@mail.xrlian.com
 \\
       \addr Department of Computer Science\\
       University of Rochester\\
       Rochester, USA
\AND
       \name 
Ji Liu \email ji.liu.uwisc@gmail.com
 \\
       \addr Ytech Seattle AI Lab, FeDA Lab, AI platform\\
       Kwai Inc.\\
       Seattle, USA}

\editor{}

\maketitle

\begin{abstract}
Stochastic compositional optimization arises in many important machine learning applications.
The objective function is the composition of two expectations of stochastic functions,
 and is more challenging to optimize than vanilla stochastic optimization problems.
In this paper, we investigate the 
stochastic compositional optimization in the general smooth non-convex setting.
We employ a recently developed idea of \textit{Stochastic Recursive Gradient Descent} to design a novel algorithm named SARAH-Compositional, and prove a sharp Incremental First-order Oracle (IFO) complexity upper bound for stochastic compositional optimization: $\mathcal{O}((n+m)^{1/2} \varepsilon^{-2})$ in the finite-sum case and $\mathcal{O}(\varepsilon^{-3})$ in the online case.
Such a complexity is known to be the best one among IFO complexity results for non-convex stochastic compositional optimization. Numerical experiments on risk-adverse portfolio management, value function evaluation in reinforcement learning, and stochastic neighborhood embedding validate the superiority of SARAH-Compositional over a few rival algorithms.
\end{abstract}

\section{Introduction}\label{sec:intro}
We consider the general smooth, non-convex compositional optimization problem of minimizing the composition of two expectations of stochastic functions:
\beq\label{eq:problem}
\min_{x \in \RR^d}~ \bigg\{  \Phi(x) \equiv (f\circ g)(x)  \bigg\}
,
\eeq
where the outer and inner functions $f : \RR^{l} \rightarrow \RR,
g: \RR^{d} \rightarrow \RR^{l}$ are defined as $f(y) := \mathbb{E}_v[f_v(y)]$,
$g(x) := \mathbb{E}_w[g_w(y)]$, $v$ and $w$ are random variables, and each component $f_v$, $g_w$ are
smooth but 
\textit{not} necessarily convex.
Compositional optimization can be used to formulate many important machine learning problems, e.g.~reinforcement learning \citep{SUTTON-BARTO}, risk management \citep{dentcheva2017statistical}, multi-stage stochastic programming \citep{SHAPIRO-DENTCHEVA-RUSZCZYNSKI}, deep neural net \citep{yang2019multilevel}, etc.
We list two specific application instances that can be written in the stochastic compositional form of \eqref{eq:problem}:

\begin{itemize}
\item
\textbf{Risk-adverse portfolio management problem}, which is formulated as

\beq\label{eq:risk_management}
\min_{x \in \RR^N}~~
 - 
 \frac{1}{T}\sum_{t=1}^T \langle r_t, x \rangle 
 + 
 \frac{1}{T} \sum_{t=1}^T \left( \langle r_t, x\rangle - \frac{1}{T} \sum_{s=1}^T \langle r_s, x\rangle \right)^2
 ,
\eeq
where $r_t\in \RR^N$ denotes the returns of $N$ assets at time $t$, and $x \in \RR^N$ denotes the investment quantity corresponding to $N$ assets. The goal is to maximize the return while 
controlling the variance of the portfolio.
\eqref{eq:risk_management} can be written as a compositional optimization problem with two functions

\begin{align}
g(x) &= 
\left[
x_1, x_2, \dots, x_N, \frac{1}{T}\sum_{s=1}^T \langle r_s, x\rangle
\right]^\top
,\\
f(w) &= 
-\frac{1}{T}\sum_{t=1}^T \langle r_t, w_{\backslash (N+1)}\rangle
 + 
\frac{1}{T}\sum_{t=1}^T\left( \langle r_t, w_{\backslash (N+1)} \rangle - w_{N+1}\right)^2
,
\label{eqRL}
\end{align}
where $w_{\backslash (N+1)}$ denotes the (column) subvector consisting of the first $N$ coordinates of $w$, and $w_{N+1}$ denotes the $(N+1)$-th coordinate of $w$.

\item
\textbf{Value function evaluation in reinforcement learning}, where the objective function of interest is
\begin{equation}
\label{bellman_1}
\EE\bigg(
	V^{\pi}(s_1) - \EE\left[ r_{s_1,s_2} + \gamma V^{\pi} (s_2) \mid s_1  \right]
\bigg)^2
,
\end{equation}
where $s_1, s_2 \in \mathcal{S}$ are two plausible states, $r_{s_1, s_2}$ denotes the reward to move from $s_1$ to $s_2$, and $V^\pi(s)$ is the value function on state $s$ corresponding to policy $\pi$.

\item
\textbf{Stochastic neighbor embedding (SNE)}, where we use $z$'s to denote points in high dimensional space, $x$'s to denote their low dimensional images and $p_{i\mid n}$ to denote the distance between $z_t$ and $z_n$ under a predefined measure. The SNE problem can be formulated as a non-convex compositional optimization problem \citep{liu2017variance} as \eqref{eq:problem} and \eqref{COOF}, where
$$
g_j(x)
=
\bigg[
x, e^{- \|x_1 - x_j\|^2} - 1, \dots, e^{- \|x_n - x_j\|^2} - 1
\bigg]^\top
,
$$
$$
f_i(w)
=
p_{i | 1} \left(
\|w_i - w_1\|^2 - \log(w_{n + 1})
\right)
+ \dots +
p_{i | n} \left(
\|w_i - w_n\|^2 - \log(w_{n + n})
\right)
.
$$
\end{itemize}

Compared with vanilla stochastic optimization problem where the optimizer is allowed to access the stochastic gradients, stochastic compositional problem~\eqref{eq:problem} is more difficult to solve.
Classical algorithms for solving \eqref{eq:problem} are often more computationally challenging.
This is mainly due to the nonlinear structure of the composition function with respect to the 
random index pair 
$(v,w)$.
Treating the objective function as an expectation $\mathbb{E}_v f_v(g(x))$, computing each iterate of the gradient estimation involves recalculating $g(x) = \mathbb{E}_w g_w(x)$, which is either time-consuming or impractical.
To tackle such weakness in practition, \citet{wang2017stochastic} firstly introduce a two-time-scale algorithm called Stochastic Compositional Gradient Descent (SCGD) along with its (in Nesterov's sense) accelerated variant (Acc-SCGD), and provide a first convergence rate analysis to that problem.
Subsequently, \citet{wang2017accelerating} proposed accelerated stochastic compositional proximal gradient algorithm (ASC-PG) which improves over the upper bound complexities in \citet{wang2017stochastic}.
Furthermore, variance-reduced gradient methods designed specifically for compositional optimization on non-convex settings arises from~\citet{liu2017variance} and later generalized to the non-smooth setting~\citep{huo2018accelerated}.
These approaches aim at getting variance-reduced estimators of $g$, $\partial g$ and $\partial g(x) \nabla f(g(x))$, respectively.
Such success signals the necessity and possibility of designing a special algorithm for non-convex objectives with better convergence rates.

\begin{table*}[t]
  \centering
\begin{tabular}{|l|c|c|c|c|}
\hline
Algorithm
&
Finite-sum
&
Online
\\ \hline\hline
SCGD \citep{wang2017stochastic}
&
unknown
&
$\varepsilon^{-8}$
\\ \hline
Acc-SCGD \citep{wang2017stochastic}
&
unknown
&
$\varepsilon^{-7}$
\\ \hline
ASC-PG \citep{wang2017accelerating}
&
unknown
&
$\varepsilon^{-4.5}$
\\ \hline
SCVR / SC-SCSG \citep{liu2017variance}
&
$(n+m)^{4/5}\varepsilon^{-2}$
&
$\varepsilon^{-3.6}$
\\ \hline
VRSC-PG \citep{huo2018accelerated}
&
$(n+m)^{2/3} \varepsilon^{-2}$
&
unknown
\\ \hline
{\bf
SARAH-Compositional (this work)
}
&
{\bf
$(n+m)^{1/2} \varepsilon^{-2}$
}
&
{\bf
$\varepsilon^{-3}$
}
\\ \hline
\end{tabular}%
\caption{Comparison of IFO complexities with different algorithms for general non-convex problem.}
\label{table}
\end{table*}


In this paper, we propose an efficient algorithm called SARAH-Compositional for the stochastic compositional optimization problem \eqref{eq:problem}. 
For notational simplicity, we let $n,m\ge 1$ and the index pair $(v,w)$ be uniformly distributed over the product set $[1,n]\times [1,m]$, i.e.
\beq\label{COOF}
\Phi(x)
=
\frac{1}{n} \sum\limits_{i=1}^n f_i\left(\frac{1}{m}\sum\limits_{j=1}^m g_j(x)\right) \ .
\eeq
We use the same notation for the online case, in which case either $n$ or $m$ can be infinite.
A fundamental theoretical question for stochastic compositional optimization is the Incremental First-order Oracle (IFO) (the number of individual gradient and function evaluations;
see Definition \ref{defi:IFO} in \S\ref{sec:algoSCGD} for a precise definition) complexity bounds for stochastic compositional optimization.
Our new SARAH-Compositional algorithm is developed by integrating the iteration of \textit{stochastic recursive gradient descent}~\citep{nguyen2017sarah}, shortened as SARAH,\footnote{This is also referred to as stochastic recursive variance reduction method, incremental variance reduction method or SPIDER-BOOST in various recent literatures. We stick to name the algorithm after SARAH to respect to our best knowledge the earliest discovery of that algorithm.} with the stochastic compositional optimization formulation~\citep{wang2017stochastic}.
The motivation of this approach is that SARAH with specific choice of stepsizes is known to be \textit{optimal} in stochastic optimization and regarded as a cutting-edge variance reduction technique, with significantly reduced oracle access complexities than earlier variance reduction method \citep{fang2018spider}.
We prove that SARAH-Compositional can reach an IFO computational complexity of $\mathcal{O}(\min\left(
(n+m)^{1/2} \varepsilon^{-2}, \varepsilon^{-3}
\right))$, improving the best known result
of $\mathcal{O}(\min\left((n+m)^{2/3} \varepsilon^{-2}, \varepsilon^{-3.6}\right) )$ in non-convex compositional optimization.
See Table~\ref{table} for detailed comparison.

\noindent\textbf{Related Works }
Classical first-order methods such as gradient descent (GD), accelerated gradient descent (AGD) and stochastic gradient descent (SGD) have received intensive attetions in both convex and non-convex optimization
~\citep{NESTEROV,ghadimi2016accelerated,li2015accelerated}. 
When the objective can be written in a finite-sum or online/expectation structure, 
variance-reduced gradient (a.k.a.~variance reduction) techniques including SAG \citep{schmidt2017minimizing}, SVRG \citep{xiao2014proximal, allen2016variance,reddi2016stochastic}, SDCA \citep{shalev2013stochastic,shalev2014accelerated}, SAGA~\citep{defazio2014saga}, SCSG~\citep{lei2017non}, SNVRG~\citep{zhou2018stochastic}, SARAH/SPIDER~\citep{nguyen2017sarah,fang2018spider,wang2019spiderboost,nguyen2019optimal}, etc., can be employed to improve the theoretical convergence properties of classical first-order algorithms. 
Notably in the smooth non-convex setting,~\citet{fang2018spider} recently proposed the SPIDER-SFO algorithm which non-trivially hybrids the iteration of stochastic recursive gradient descent (SARAH)~\citep{nguyen2017sarah} with the normalized gradient descent. 
In the representative case of batch-size 1, SPIDER-SFO adopts a small step-length that is proportional to $\varepsilon^2 \land \varepsilon n^{-1/2}$ where $\varepsilon$ is the squared targeted accuracy, and (by rebooting the SPIDER tracking iteration once every $n \land \cO(\varepsilon^{-2})$ iterates) the variance of the stochastic estimator can be constantly controlled by $\cO(\varepsilon^2)$. 
For finding $\varepsilon$-accurate solution purposes,~recent works \citet{wang2019spiderboost,nguyen2019optimal} discovered two variants of the SARAH algorithm that achieve the same complexity as SPIDER-SFO~\citep{fang2018spider} and SNVRG~\citep{zhou2018stochastic}.\footnote{\citet{wang2019spiderboost} names their algorithm SPIDER-BOOST since it can be seen as the SPIDER-SFO algorithm with relaxed step-length restrictions.} The theoretical convergence property of SARAH/SPIDER methods in the smooth non-convex case outperforms that of SVRG, and is provably optimal under a set of mild assumptions~\citep{arjevani2019lower,fang2018spider,nguyen2019optimal,wang2019spiderboost}.

It turns out that when solving compositional optimization problem~\eqref{eq:problem}, classical first-order methods for optimizing a single objective function can either be non-applicable or it brings at least $\cO(m)$ queries to calculate the inner function $g$.
To remedy this issue,~\citet{wang2017stochastic,wang2017accelerating} considered the stochastic setting and proposed the SCGD algorithm to calculate or estimate the inner finite-sum more efficiently, achieving a polynomial rate that is independent of $m$.
Later on, \citet{lian2017finite,liu2017variance,huo2018accelerated} and \citet{lin2018improved} merged SVRG method into the compositional optimization framework to do variance reduction on all three steps of the estimation.
In stark contrast, our work adopts the SARAH/SPIDER method which is theoretically more efficient than the SVRG method in the non-convex compositional optimization setting.

After the initial submission of the short version of this technical report, we are aware of a line of concurrent works by Zhang and Xiao \citep{zhang2019stochastic,zhang2019multi} who adopted the idea of SPIDER \citet{fang2018spider} and solve the stochastic compositional problem.
More relevant to this work is \citet{zhang2019multi} which consider a special non-smooth setting for the compositional optimization problem where the objective function has an additive non-smooth term that admits an easy proximal mapping.\footnote{Such a setting has also been studied in \citet{wang2017accelerating,lin2018improved,huo2018accelerated}, among many others.}
We omit the non-smooth part for a fair comparison, and the IFO complexity upper bound obtained in \citet{zhang2019multi} is similar to ours (Theorems \ref{theo:SFOSPfinitesum} and \ref{theo:SFOSPonline}).
There are two significant differences between the two lines of works:
(i)
\citet{zhang2019multi} suffers from the step-length restriction that SPIDER has in nature, and our work circumvent this issue, hence applicable to a wider range of statistical learning tasks;
(ii)
Our work theoretically optimizes the choice of batch sizes (Corollary \ref{coro:OPTBSfinitesum} and contexts) and further halves the IFO upper bound in the asymptotic regime $1 \ll 2m+n \ll \ep^{-4}$ (\citet{zhang2019multi} fixes the batch size parameters in their comparable result), which potentially serves as a parameter-tuning guidance to practitioners.
\citet{zhang2019stochastic,zhang2019multi} also study other important cases including adaptive batch size and multilevel nested compositional optimization \citep{yang2019multilevel} and obtain sharp convergence rates.

\noindent\textbf{Contributions }
This work makes two contributions as follows.
First, we propose a new algorithm for stochastic compositional optimization called SARAH-Compositional, which operates SARAH/SPIDER-type recursive variance reduction to estimate relevant quantities.
Second, we conduct theoretical analysis for both online and finite-sum cases, which verifies the superiority of SARAH-Compositional over the best known previous results.
In the finite-sum case, we obtain a complexity of $(n+m)^{1/2} \varepsilon^{-2}$ which improves over the best known complexity $(n+m)^{2/3} \varepsilon^{-2}$ achieved by \citet{huo2018accelerated}.
In the online case we obtain a complexity of $\varepsilon^{-3}$ which improves the best known complexity $\varepsilon^{-3.6}$ obtained in \citet{liu2017variance}.

\noindent\textbf{Notational Conventions }
Throughout the paper, we treat the parameters $L_g, L_f, L_\Phi, M_g, M_f, \Delta$ and $\sigma$ as global constants.
Let $\|\bullet\|$ denote the Euclidean norm of a vector or the operator norm of a matrix induced by Euclidean norm, and let $\|\bullet\|_F$ denotes the Frobenious norm. 
For fixed $T\geq t\geq 0$ let $x_{t:T}$ denote the sequence $\{x_t,...,x_T\}$. Let $\mathbf{E}_t[\bullet]$ denote the conditional expectation
$\mathbf{E}[\bullet|x_0,x_1,...,x_t]$.
Let $[1,n]=\{1,...,n\}$ and $S$ denote the cardinality of a multi-set $\mathcal{S}\subseteq [1,n]$ of samples (a generic set that permits repeated instances).
The averaged sub-sampled stochastic estimator is denoted as $\mathcal{A}_\mathcal{S}=(1/S)\sum\limits_{i\in \mathcal{S}}\mathcal{A}_i$ where the summation counts repeated instances. 
We denote $p_n=\mathcal{O}(q_n)$ if there exist some constants $0<c<C<\infty$ such that $cq_n\leq p_n\leq Cq_n$ as $n$ becomes large.
Other notations are explained at their first appearances.

\noindent\textbf{Organization }
The rest of our paper is organized as follows.
\S\ref{sec:algoSCGD} formally poses our algorithm and assumptions.
\S\ref{Section:Analysis-SARAH-SCGD} presents the convergence rate theorem and \S\ref{sec:experiments} presents numerical experiments 
that apply our algorithm to the task of portfolio management. 
We conclude our paper in \S\ref{sec:conclusion}.
Proofs of convergence results for finite-sum and online cases and auxiliary lemmas are deferred to \S\ref{sec:app_proof} and \S\ref{sec:app_aux} in the supplementary material.

\section{SARAH for Stochastic Compositional Optimization} \label{sec:algoSCGD}
Recall our goal is to solve the compositional optimization problem \eqref{eq:problem}, i.e. to minimize $\Phi(x) = f(g(x))$ where
$$
f(y) := \frac{1}{n}\sum_{i=1}^n f_i(y)
 ,\qquad
g(x) := \frac{1}{m} \sum_{j=1}^m g_j(x)
.
$$
Here for each $j\in [1,m]$ and $i\in [1,n]$ the functions $g_j: \RR^d\rightarrow \RR^l$ and $f_i: \RR^l\rightarrow \RR$.
We can formally take the derivative of the function $\Phi(x)$ and obtain (via the chain rule) the gradient descent iteration

\begin{equation}\label{Eq:COGD}
x_{t+1}=x_t-\eta  [\partial g(x_t)]^\top \nabla f(g(x_t)) \ ,
\end{equation}
where the $\partial$ operator computes the Jacobian matrix of the smooth mapping, and the gradient operator $\nabla$ is only taken with respect to the first-level variable. 
As discussed in \S\ref{sec:intro}, it can be either impossible (online case) or time-consuming (finite-sum case) to estimate the terms $\displaystyle{\partial g(x_t)=\dfrac{1}{m}\sum\limits_{j=1}^m \partial g_j(x_t)}$ and $\displaystyle{g(x_t)=\dfrac{1}{m}\sum\limits_{j=1}^m g_j(x_t)}$ in the iteration scheme~\eqref{Eq:COGD}. 
In this paper, we design a novel algorithm (SARAH-Compositional) based on Stochastic Compositional Variance Reduced Gradient method (see \citet{lin2018improved}) yet hybriding with the stochastic recursive gradient method \citet{nguyen2017sarah}.
As the readers see later, our SARAH-Compositional is more efficient than all existing algorithms for non-convex compositional optimization.

We introduce some definitions and assumptions.
First, we assume the algorithm has accesses to an incremental first-order oracle in our black-box environment \citep{lin2018improved};
 also see \citep{agarwal2015lower,woodworth2016tight} for vanilla optimization case:

\begin{definition}[IFO]\label{defi:IFO}\citep{lin2018improved}
The Incremental First-order Oracle (IFO) returns, when some $x \in \RR^d$ and $j \in [1,m]$ are inputted, the vector-matrix pair $[g_j(x), \partial g_j(x)]$ or when some $y \in \RR^l$ and $i \in [1,n]$ are inputted, the scalar-vector pair $[f_i(y), \nabla f_i(y)]$.
\end{definition}

Second, our goal in this work is to find an $\varepsilon$-accurate solution, defined as

\begin{definition}[$\varepsilon$-accurate solution]\label{defi:FOSP}
We call $x\in \RR^d$ an $\varepsilon$-accurate solution to problem \eqref{eq:problem}, if
\beq\label{defi:FOSP_ASP}
  \|\nabla \Phi(x)\|\le \varepsilon.
\eeq
\end{definition}
It is worth remarking here that the inequality \eqref{defi:FOSP_ASP} can be
modified to $\|\nabla \Phi(x)\|\leq C\varepsilon$ for some \textit{global} constant $C>0$ without hurting the magnitude of IFO complexity bounds.

Let us first make some assumptions regarding to each component of the (compositional) objective function.
Analogous to Assumption 1(i) of \citet{fang2018spider}, we make the following finite gap assumption:

\begin{assumption}[Finite gap]\label{Assumption:GapMinimizer}
We assume that the algorithm is initialized at $x_0\in \RR^d$ with
\begin{equation}\label{Eq:SARAH-SCGD:Assumption:GapMinimizer}
\Delta:=\Phi(x_0)-\Phi^\ast<\infty \ ,
\end{equation}
where $\Phi^\ast$ denotes the global minimum value of $\Phi(x)$.
\end{assumption}

We make the following standard smoothness and boundedness assumptions, which are standard in recent compositional optimizatioin literatures (e.g.~\citet{lian2017finite,huo2018accelerated,lin2018improved}).

\begin{assumption}[Smoothness]\label{Assumption:Smoothness}
There exist Lipschitz constants $L_g, L_f, L_\Phi >0$ such that for $i\in [1,n]$, $j\in [1,m]$ we have

\beq\label{assmoothness}
\begin{aligned}
&\|\partial g_j(x)-\partial g_j(x')\|_F &\le L_g \|x-x'\|
&&\quad
\text{for } x,x'\in \RR^d
, \\
&\|\nabla f_i(y)-\nabla f_i(y')\| &\le L_f \|y-y'\|
&&\quad
\text{for } y,y'\in \RR^l
, \\
&\left\|[\partial g_j(x)]^\top\nabla f_i(g(x))-[\partial g_j(x')]^\top\nabla f_i(g(x'))\right\| &\le L_\Phi \|x-x'\|
&&\quad
\text{for } x,x'\in \RR^d
.
\end{aligned}\eeq

\end{assumption}
Here for the purpose of using stochastic recursive estimation of $\partial g(x)$, we slightly strengthen the smoothness assumption by adopting the Frobenius norm in left hand of the first line of \eqref{assmoothness}.

\begin{assumption}[Boundedness]\label{Assumption:Boundedness}
There exist boundedness constants $M_g, M_f>0$ such that for $i\in [1,n]$, $j\in [1,m]$ we have
\beq\label{asbounded}
\begin{aligned}
&\|\partial g_j(x)\| \le M_g
&&\quad
\text{for } x\in \RR^d
,\\
&\|\nabla f_i(y)\| \le M_f
&&\quad
\text{for } y\in \RR^l
.
\end{aligned}\eeq
\end{assumption}
Notice that applying the fundamental theorem of calculus to \eqref{asbounded}  gives another Lipschitz condition
\begin{equation}\label{gsmooth}
\|g_j(x)-g_j(x')\| \leq M_g \|x-x'\|
\qquad
\text{ for } x,x'\in \RR^d
\ ,
\end{equation}
and analogously for $ f_i(y)$.
It turns out that under the above two assumptions, a choice of $L_\Phi$ in \eqref{assmoothness} can be expressed as a polynomial of $L_f, L_g, M_f, M_g$.
For clarity purposes in the rest of this paper, 
we adopt the following typical choice of $L_\Phi$:
\beq\label{LPhi}
L_\Phi \equiv M_f L_g + M_g^2 L_f
.
\eeq
whose applicability can be verified via a simple application of the chain rule.
We integrate both finite-sum and online cases into one algorithm SARAH-Compositional and write it in Algorithm \ref{Alg:SARAH-SCGD}.

\begin{algorithm}[t]
\caption{SARAH-Compositional, Online Case (resp.~Finite-Sum Case)}
\label{Alg:SARAH-SCGD}
\begin{algorithmic}
\STATE \textbf{Input}: $T, q, x_0, \eta, S^L_1, S_1, S^L_2, S_2, S^L_3, S_3$
\FOR {$t=0$ \TO $T-1$}
    \IF {$\text{mod}(t,q)=0$}
\STATE
Draw $S^L_1$ samples and let $\boldsymbol{g}_t= \dfrac{1}{S^L_1} \sum\limits_{j\in \cS^L_{1,t}} g_j(x_t)$
(resp.~$\bg_{t}=g\left(x_{t}\right)$ in finite-sum case)
\STATE
Draw $S^L_2$ samples and let $\boldsymbol{G}_t= \dfrac{1}{S^L_2} \sum\limits_{j\in \cS^L_{2,t}} \partial g_j(x_t)$
(resp.~$\bG_{t}=\partial g\left(x_{t}\right)$ in finite-sum case)
\STATE
Draw $S^L_3$ samples and let $\boldsymbol{F}_t= \dfrac{1}{S^L_3} \left(\bG_{t}\right)^{\top} \sum\limits_{i\in \cS^L_{3,t}} \nabla f_{i}(\boldsymbol{g}_t)$
(resp.~$\bF_{t}=\left(\bG_{t}\right)^{\top} \nabla f\left(\bg_{t}\right)$ in finite-sum case)

    \ELSE
\STATE
Draw $S_1$ samples and let
$\boldsymbol{g}_t= \dfrac{1}{S_1} \sum\limits_{j\in \cS_{1,t}} g_j(x_t) - \dfrac{1}{S_1} \sum\limits_{j\in \cS_{1,t}} g_j(x_{t-1}) + \boldsymbol{g}_{t-1}$

\STATE
Draw $S_2$ samples and let
$\boldsymbol{G}_t=\dfrac{1}{S_2} \sum\limits_{j\in \cS_{2,t}} \partial g_j(x_t) - \dfrac{1}{S_2} \sum\limits_{j\in \cS_{2,t}} \partial g_j(x_{t-1}) + \boldsymbol{G}_{t-1}$

\STATE
Draw $S_3$ samples and let
$$\bF_{t}
  =
\frac{1}{S_3}\left(\bG_{t}\right)^{\top}\sum_{i \in \cS_{3,t}} \nabla f_i\left(\bg_{t}\right)-\frac{1}{S_3}\left(\bG_{t-1}\right)^{\top} \sum_{i \in \cS_{3,t}}\nabla f_i\left(\bg_{t-1}\right)
+
\bF_{t-1}$$

    \ENDIF
\STATE
Update $x_{t+1}=x_{t} - \eta \bF_{t}$
\ENDFOR
\RETURN Output $\widetilde{x}$ chosen uniformly at random from $\{x_t\}_{t=0}^{T-1}$
\end{algorithmic}
\end{algorithm}

\section{Convergence Rate Analysis}
\label{Section:Analysis-SARAH-SCGD}
In this section, we aim to justify that our proposed SARAH-Compositional algorithm provides IFO complexities of $\mathcal{O}\left((n+m)^{1/2} \varepsilon^{-2}\right)$ in the finite-sum case and $\mathcal{O}(\varepsilon^{-3})$ in the online case, which supersedes the concurrent and comparative algorithms (see more in Table \ref{table}).

Let us first analyze the convergence in the finite-sum case.
In this case we have $\mathcal{A}_1=[1,m]$, $\mathcal{B}_1=[1,m]$, $\mathcal{C}_1=[1,n]$. Involved analysis leads us to conclude
\begin{theorem}[Finite-sum case]\label{theo:SFOSPfinitesum}
Suppose Assumptions \ref{Assumption:GapMinimizer}, \ref{Assumption:Smoothness} and \ref{Assumption:Boundedness} in \S\ref{sec:algoSCGD} hold,
let $\mathcal{A}_1 = \mathcal{B}_1 = [1,m]$, $\mathcal{C}_1 = [1,n]$,
let for any mini-batch sizes $S_1, S_2 \in [1,m]$, $S_3 \in [1,n]$
\beq\label{Soqeq}
S_o
 = 
\left(1 + \dfrac{2}{S_3} \right)  \left(\dfrac{M_g^4L_f^2}{S_1} + \dfrac{M_f^2 L_g^2}{S_2}\right)
,
\qquad
q =  \frac{2m+n}{S_1 + S_2 + S_3} 
,
\eeq
and set the stepsize
\beq\label{etace}
\eta    =  \frac{1}{ \max\left( \sqrt{6S_o q},  2L_\Phi \right)}
.
\eeq
Then for the finite-sum case, SARAH-Compositional Algorithm \ref{Alg:SARAH-SCGD} outputs an $\widetilde{x}$ satisfying $\mathbf{E}\|\nabla \Phi(\widetilde{x})\|^2\leq \varepsilon^2$ in
\beq\label{Qiter}
     \frac{2[ \Phi(x_0) - \Phi^\ast] }{\varepsilon^2} \cdot
     \max\left( \sqrt{6\left(1 + \frac{2}{S_3}\right) \left(\frac{M_g^4L_f^2}{S_1} + \frac{M_f^2 L_g^2}{S_2}\right) \left(\frac{2m+n}{S_1 + S_2 + S_3} \right)},  2L_\Phi \right)
\eeq
iterates.
Furthermore, let the mini-batch sizes $S_1, S_2 \in [1,m]$, $S_3 \in [1,n]$ satisfy
\beq\label{bscond}
3\left(1 + \frac{2}{S_3}\right) \left(\frac{M_g^4L_f^2}{S_1} + \frac{M_f^2 L_g^2}{S_2}\right) \left(
 \frac{2m+n}{S_1 + S_2 + S_3} 
\right)
 \ge 
2L_\Phi^2
,
\eeq
then the IFO complexity to achieve an $\varepsilon$-accurate solution is bounded by
\beq\label{IFObound}
2m+n
 + 
\sqrt{2m + n}
 \cdot
\sqrt{
   (S_1+S_2+S_3 )   
  \left(1 + \frac{2}{S_3}\right) \left(\frac{M_g^4L_f^2}{S_1} + \frac{M_f^2 L_g^2}{S_2}\right)
}
\cdot
  \frac{\sqrt{216} [ \Phi(x_0) - \Phi^\ast] }{\varepsilon^2}
.
\eeq
\end{theorem}

Like in \citet{fang2018spider}, for a wide range of mini-batch sizes the IFO complexity to achieve an $\varepsilon$-accurate solution is upper bounded by $\cO((m+n)^{1/2} \varepsilon^{-2})$, as long as \eqref{bscond} holds.%
\footnote{%
Here and in below, the smoothness and boundedness parameters and $\Phi(x_0) - \Phi^\ast$ are treated as constants%
}
Note if the batch size are chosen as $S_1 = S_2 = S_3 = 1$, then from \eqref{IFObound} the IFO complexity upper bound is
\beq\label{IFOsingle}
2m+n
 + 
\sqrt{2m + n} \cdot
\sqrt{ 9 \left(M_g^4 L_f^2  + M_f^2 L_g^2 \right)} \cdot  \frac{\sqrt{216} [ \Phi(x_0) - \Phi^\ast] }{\varepsilon^2}
.
\eeq

Let us then analyze the convergence in the online case, where we sample mini-batches $\mathcal{A}_1, \mathcal{B}_1, \mathcal{C}_1$ of relevant quantities instead of the ground truth once every $q$ iterates.
To characterize the estimation error, we put in one additional finite variance assumption:
\begin{assumption}[Finite Variance]
\label{assump:finite_variance}
We assume that there exists $H_1, H_2$ and $H_3$ as the upper bounds on the variance of the functions $f(y)$, $\partial g(x)$, and $g(x)$, respectively, such that

\beq
\label{assumpeq:last}
\begin{aligned}
&\EE\|\nabla f_{i}(y) - \nabla f(y)\|^2 \leq H_1
& \text{for}\quad y \in \RR^l
,\\
&\EE\|\partial g_{i}(x) - \partial g(x)\|^2 \leq H_2
& \text{for}\quad x \in \RR^d
,\\
&\EE\|g_{i}(x) - g(x)\|^2 \leq H_3
& \text{for}\quad x \in \RR^d
.
\end{aligned}
\eeq
\end{assumption}
From Assumptions \ref{Assumption:Smoothness} and \ref{Assumption:Boundedness} we can easily verify, via triangle inequality and convexity of norm, that $H_2$ can be chosen as $4M_g^2$ and $H_1$ can be chosen as $4M_f^2$.
On the contrary, $H_3$ \textit{cannot} be represented as a function of boundedness and smoothness constants.
We conclude the following theorem for the online case:

\begin{theorem}[Online case]\label{theo:SFOSPonline}
Suppose Assumptions \ref{Assumption:GapMinimizer}, \ref{Assumption:Smoothness} and \ref{Assumption:Boundedness} in \S\ref{sec:algoSCGD} hold,
let $S^L_1 =\dfrac{3H_3 M_g^2L_f^2}{\varepsilon^2}, S^L_2 = \dfrac{3H_2 M_f^2}{\varepsilon^2}$, $S^L_3 = \dfrac{3H_1 M_g^2}{ \varepsilon^2}$,
let for any mini-batch sizes $S_1 ,  S_2, S_3 $
\beq\label{Soqeq2}
S_o
 = 
\left(1 + \dfrac{2}{S_3} \right)  \left(\dfrac{M_g^4L_f^2}{S_1} + \dfrac{M_f^2 L_g^2}{S_2}\right),
\eeq
let noise-relevant parameter
\beq\label{D_0}
D_0 := 3H_3M_g^2L_f^2 + 3H_2M_f^2 + 3H_1M_g^2
,
\eeq
let $q =  \dfrac{D_0}{\varepsilon^2 (S_1 + S_2 + S_3)}$, and set the stepsize
\beq\label{etace2}
\eta    =  \frac{1}{ \max\left( \sqrt{6S_o q},  2L_\Phi \right)}
.
\eeq
Then the SARAH-Compositional Algorithm \ref{Alg:SARAH-SCGD} outputs an $\widetilde{x}$ satisfying $\mathbf{E}\|\nabla \Phi(\widetilde{x})\|^2\leq 2\varepsilon^2$ in
\beq\label{Qiter2}
     \frac{2[ \Phi(x_0) - \Phi^\ast] }{\varepsilon^2} \cdot
     \max\left( \sqrt{6D_0\left(1 + \frac{2}{S_3}\right) \left(\frac{M_g^4L_f^2}{S_1} + \frac{M_f^2 L_g^2}{S_2}\right) \frac{1}{ (S_1 + S_2 + S_3)\varepsilon^2}},  2 L_\Phi \right)
\eeq
iterates.
Furthermore, let the mini-batch sizes $S_1, S_2$, $S_3$ satisfy
\beq\label{bscond2}
3\left(1 + \frac{2}{S_3}\right) \left(\frac{M_g^4L_f^2}{S_1} + \frac{M_f^2 L_g^2}{S_2}\right) \frac{D_0}{(S_1 + S_2 + S_3)} 
 \ge 
2L_\Phi^2 \varepsilon^2
,
\eeq
then the IFO complexity to achieve an $\varepsilon$-accurate solution is bounded by
\beq\label{IFObound2}
\frac{D_0}{\varepsilon^2}
 + 
\sqrt{D_0} 
 \cdot 
\sqrt{(S_1 + S_2 + S_3)\left(1 + \dfrac{2}{S_3} \right)  \left(\dfrac{M_g^4L_f^2}{S_1} + \dfrac{M_f^2 L_g^2}{S_2}\right)}\cdot \frac{\sqrt{216}[ \Phi(x_0) - \Phi^\ast] }{\varepsilon^3},
\eeq

\end{theorem}

We see that in the online case, the IFO complexity to achieve an $\varepsilon$-accurate solution is upper bounded by $\cO( \varepsilon^{-3})$, as long as \eqref{bscond2} holds.%
\footnote{%
Here and in below, the smoothness and boundedness parameters and $\Phi(x_0) - \Phi^\ast$ are treated as constants%
}
Note if the batch size are chosen as $S_1 = S_2 = S_3 = 1$, then from \eqref{IFObound2} the IFO complexity upper bound is
\beq\label{IFOsingle2}
\frac{D_0}{\varepsilon^2}
 + 
\sqrt{D_0} 
 \cdot 
\sqrt{M_g^4L_f^2 + M_f^2 L_g^2}\cdot \frac{\sqrt{1944}[ \Phi(x_0) - \Phi^\ast] }{\varepsilon^3}
.
\eeq
In fact, we can further improve the coefficient in the $\varepsilon^{-2}$ term in \eqref{IFObound} and $\varepsilon^{-3}$ term in \eqref{IFObound2}.
A simple optimization tricks enables us to obtain an optimal choice (as in \eqref{bcsize} and \eqref{bcsize2} below) of mini-batch sizes, as

\begin{corollary}[Optimal batch size, finite-sum and online case]\label{coro:OPTBSfinitesum}
Let $\Pi_{[1,m]}(\cdot)$ (resp.~$\Pi_{[1,n]}(\cdot)$) maps a real to its closest element in $[1,m]$ (resp.~$[1,n]$).

\begin{itemize}
\item[(i)]
When the mini-batch sizes $S_1, S_2, S_3$ in the finite-sum case are chosen as
\beq\label{bcsize}
\begin{aligned}
&S_1
=
\Pi_{[1,m]}\left(
\frac{M_g^2 L_f}{2L_\Phi} \sqrt{6(2m+n)}
\right)
 ,\quad
\\&
S_2
=
\Pi_{[1,m]}\left(
\frac{M_f L_g}{2L_\Phi} \sqrt{6(2m+n)}
\right)
,\quad\\&
S_3
=
\Pi_{[1,n]}\left(
\sqrt[4]{6(2m+n)}
\right)
,
\end{aligned}
\eeq
the IFO complexity bound to achieve an $\varepsilon$-accurate solution for SARAH-Compositional is further minimized to
\beq\label{IFObound_optimal}
2m+n
 + 
\sqrt{2m + n}
 \cdot
\left( 1 + \sqrt[4]{\frac{8}{3(2m+n)}}  \right) (M_g^2 L_f  + M_f L_g)
\cdot
  \frac{\sqrt{216} [ \Phi(x_0) - \Phi^\ast] }{\varepsilon^2}
.
\eeq
\item[(ii)] When the mini-batch sizes $S_1, S_2, S_3$ in the online case are chosen to satisfy
\beq\label{bcsize2}
S_1
=
\frac{M_g^2 L_f}{2L_\Phi \varepsilon} \sqrt{6D_0}
 ,\qquad
S_2
=
\frac{M_f L_g}{2L_\Phi \varepsilon} \sqrt{6D_0}
,\qquad
S_3
=
\sqrt[4]{\frac{6D_0}{\varepsilon^2}}
,
\eeq
where $D_0$ is defined in~\eqref{D_0}, 
the IFO complexity bound to achieve an $\varepsilon$-accurate solution for SARAH-Compositional is further minimized to
\beq\label{IFObound_optimal2}
\frac{D_0}{\varepsilon^2}
 + 
\sqrt{D_0}
 \cdot
\left( 1 + \sqrt[4]{\frac{8}{3D_0}}  \right) (M_g^2 L_f  + M_f L_g)
\cdot
  \frac{\sqrt{216} [ \Phi(x_0) - \Phi^\ast] }{\varepsilon^3}
.
\eeq
\end{itemize}

\end{corollary}

To understand the new IFO complexity upper bounds \eqref{IFObound_optimal} and \eqref{IFObound_optimal2} with optimally chosen batch sizes, via the basic inequality
$
M_f L_g + M_g^2 L_f
 \le
\sqrt{2(M_f^2 L_g^2 + M_g^4 L_f^2 )}
,
$
the complexity in \eqref{IFObound_optimal} when $2m+n \to \infty$ can be further upper bounded by
\begin{align*}
  & \le
2m+n
 + 
\sqrt{2m + n}
 \cdot
\left( 1 + O\left((2m+n)^{-1/4}\right)  \right) 
\sqrt{2(M_f^2 L_g^2 + M_g^4 L_f^2 )}
\cdot
  \frac{\sqrt{216} [ \Phi(x_0) - \Phi^\ast] }{\varepsilon^2}
.
\end{align*}
This indicates that compared to the single-sample case \eqref{IFOsingle}, the IFO complexity upper bound obtained is reduced by at least $1 - \sqrt{2/9} = 52.86\%$ in its $\cO(\varepsilon^{-2})$ coefficient when $2m+n$ is asymptotically large.
To our best knowledge, the theoretical phenomenon that mini-batch SARAH can reduce IFO complexity has \textit{not} been quantitatively characterized in previous literatures.
It is worth noting that analogous property does \textit{not} hold in the classical optimization case, where the single-sample case and mini-batch cases share the same IFO complexity upper bound \citep{fang2018spider}.
With further efforts, it can be shown that the running time can be effectively more reduced by adopting parallel computing techniques;
we omit the details for clarity.

Due to space limits, the detailed proofs of Theorems \ref{theo:SFOSPfinitesum} and \ref{theo:SFOSPonline} and Corollary \ref{coro:OPTBSfinitesum} are deferred to \S\ref{sec:app_proof} in the supplementary material.

\section{Experiments}\label{sec:experiments}
In this section, we conduct numerical experiments to support our theory by applying our proposed SARAH-Compositional algorithm to three practical tasks: portfolio management, reinforcement learning, and a dimension reduction technique named \textit{stochastic neighborhood embedding} (SNE).
In sequel, \S\ref{sec:exp_port} studies performance of our algorithm to (risk-adverse) portfolio management/optimization problem, \S\ref{sec:exp_RL} tests the performance of SARAH-Compositional on evaluating value functions in reinforcement learning, while \S\ref{sec:tsne} focuses on the study of SNE which possesses a non-convex objective function.
We follow the setups in \citet{huo2018accelerated,liu2017variance} and compare with existing algorithms for compositional optimization.
Readers are referred to \citet{wang2017stochastic,wang2017accelerating} for more tasks we can apply our algorithm to.%
\footnote{We conduct experiments on synthetic data, real world datasets as described below and MNIST dataset; the source code can be found at \texttt{http://github.com/angeoz/SCGD}.}

\subsection{SARAH-Compositional Applied to Portfolio Management}\label{sec:exp_port}
\begin{figure*}
\hspace{-0.3in}
\centering
    \includegraphics[width=6.5in]{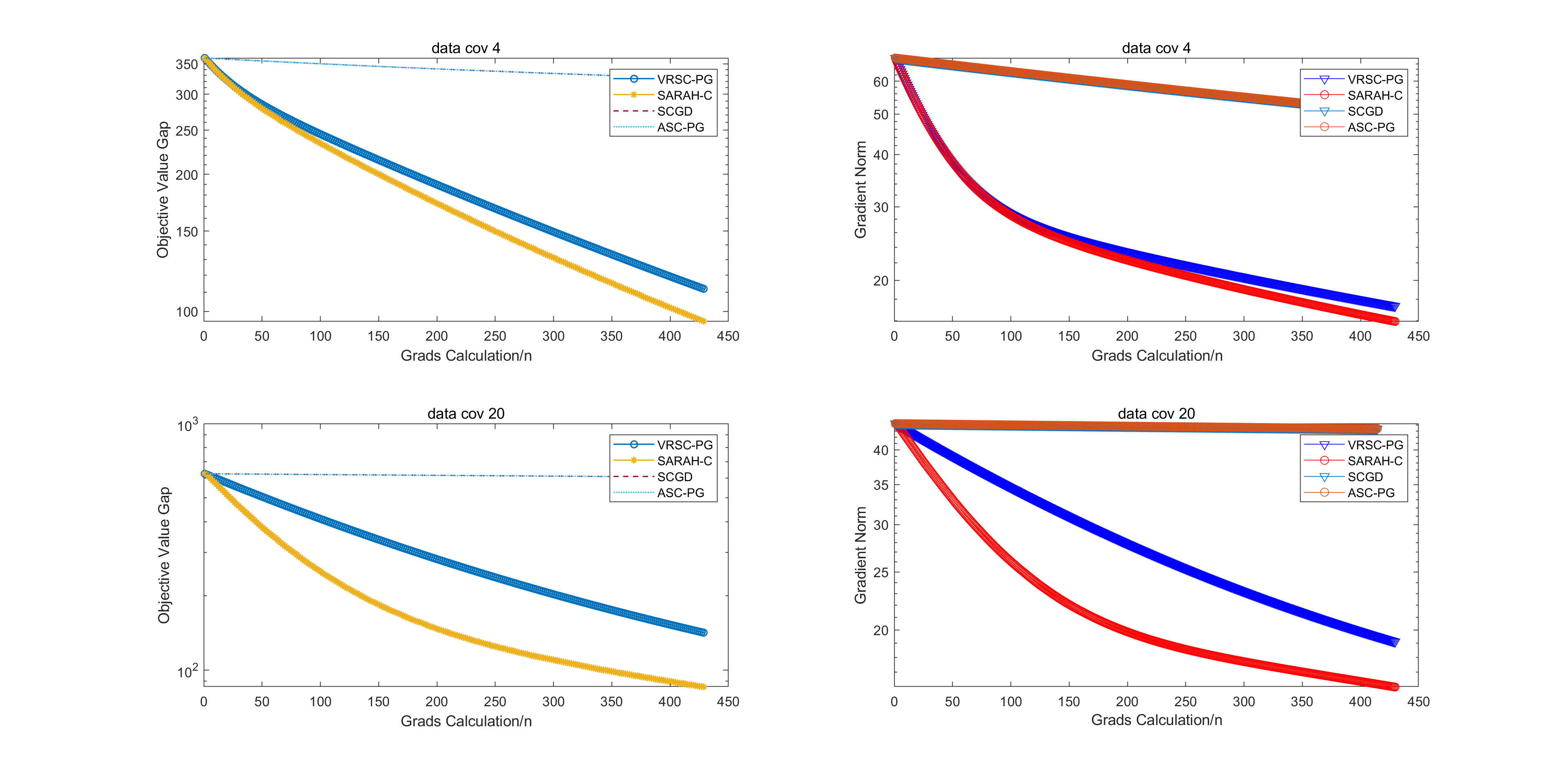}
  \caption{Experiment on the portfolio management. The $x$-axis is number of passes of the dataset, that is, the number of gradients calculations divided by the number of samples. The $y$-axis is the function value gap and the norm of gradient respectively.}
\label{fig:portfolio}
\end{figure*}

\begin{figure*}
\hspace{-0.3in}
\centering
    \includegraphics[width=6.5in]{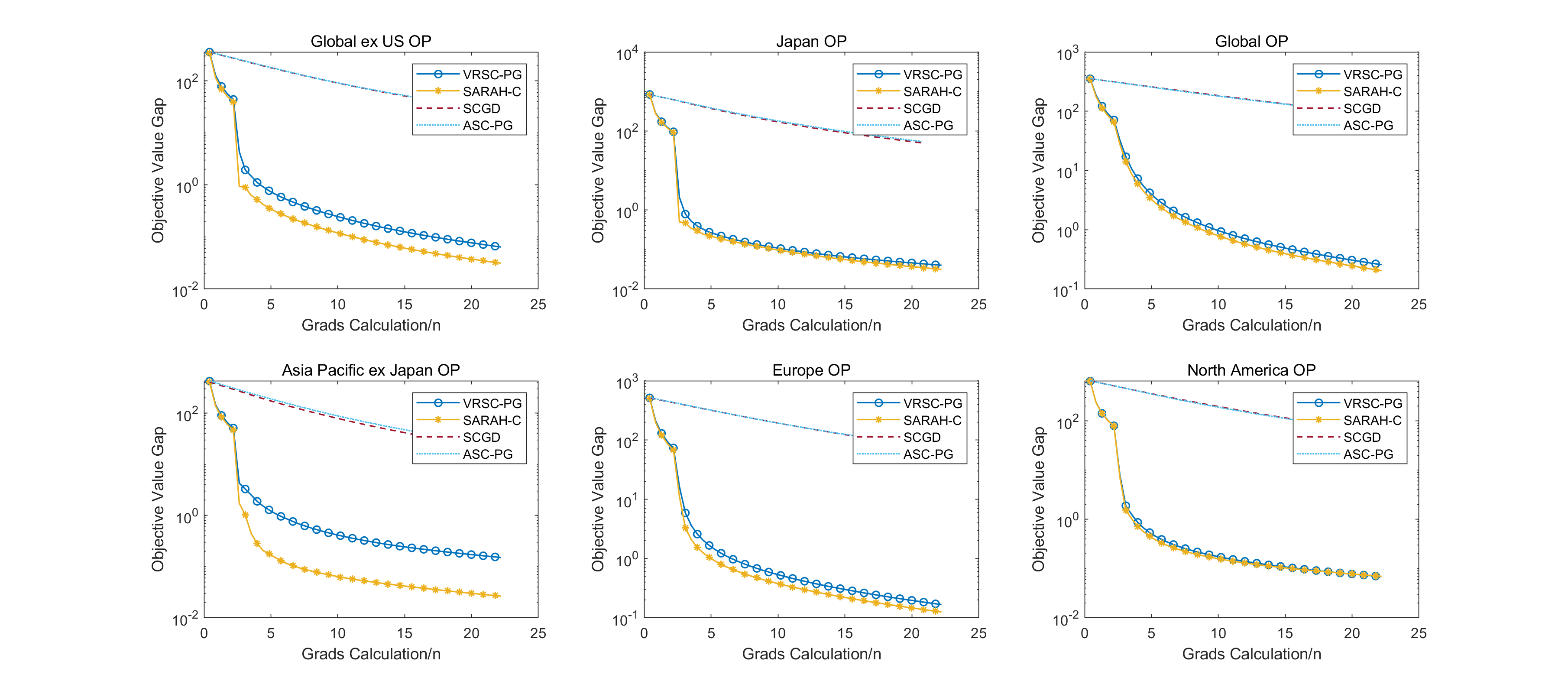}
  \caption{Experiment on the portfolio management. The $x$-axis is the number of gradients calculations divided by the number of samples, the $y$-axis is the function value gap.}
\label{fig:portfolio2}
\end{figure*}

Recall that in \S\ref{sec:intro}, we formulate our portfolio management problem as a mean-variance optimization problem \eqref{eq:risk_management}, which can be formulated as a compositional optimization problem~\eqref{eq:problem}.
As it satisfies Assumptions \ref{Assumption:GapMinimizer}--\ref{assump:finite_variance} in a bounded domain of optimization, it serves as a good example to validate our theory.
For convenience we repeat the display here:
\beq\tag{\ref{eq:risk_management}}
\min_{x \in \mathbb{R}^N}~~
 - 
 \frac{1}{T}\sum_{t=1}^T \langle r_t, x \rangle 
 + 
 \frac{1}{T} \sum_{t=1}^T \left( \langle r_t, x\rangle - \frac{1}{T} \sum_{s=1}^T \langle r_s, x\rangle \right)^2
 ,
\eeq
where $x = \left\{x_1, x_2, \ldots, x_N\right\}\in \mathbb{R}^N$ denotes the quantities invested at every asset $i=1,\dots, N$.

For illustration purpose we set $T = 2000$ and $N = 200$. When applying SARAH-Compositional we first adopt the finite-sum case i.e.~$S_1^L = S_2^L = S_3^L = T$.
Similar to the setup of \citet{huo2018accelerated}, we sample $r_t$ from the Gaussian distribution and take the absolute value.
Each row $r_t$ in the $T \times N$ matrix $[r_1, r_2, \ldots, r_T]^\top$ is (independently) generated from a zero-mean $N$-dimensional Gaussian distribution with covariance matrix $\Sigma \in \mathbb{R}^{N \times N}$.
We prescribe the conditional number $\kappa$ of the population covariance $\Sigma$ as one of our parameters, and tested the cases where $\kappa(\Sigma) = 4$ and $\kappa(\Sigma) = 20$.

When applying SARAH-Compositional to the online case where we pick $S^L_1, S^L_2, S^L_3$ as the mini-batch sizes once every $q$ steps.
Datasets include different portfolio datas formed on Size and Operating Profitability.\footnote{\url{http://mba.tuck.dartmouth.edu/pages/faculty/ken.french/data_library.html}}
We choose to use 6 different 25-portfolio datasets where $N=25$ and $T=7240$, same as the ones adopted by~\citet{lin2018improved}.
Specifically, we choose $S^L_1 = S^L_2 = S^L_3 = 2000$ (roughly optimized to improve the numerical performance).
The results are shown in Figure~\ref{fig:portfolio2}.

Throughout the experiment of the portfolio management, our search of stepsize is among $\left\{1\times 10^{-5}, 1\times10^{-4}, 2\times 10^{-4}, 5\times 10^{-4}, 1\times 10^{-3}, 1\times 10^{-2}\right\}$. The other parameters are set as follows: $q = 20$ in the finite-sum case and $q = 50$ in the online case, $S_1 =5, S_2= 5$, and $S_3= 1$. For SCGD and ASC-PG algorithm, we fix the extrapolation parameter $\beta$ to be 0.9. We plot the learning curve of each algorithms corresponding to the best learning rate found. The results are shown in Figure~\ref{fig:portfolio}and~\ref{fig:portfolio2} respectively.

We demonstrate the comparison among our algorithm SARAH-Compositional, SCGD~\citep{wang2017stochastic}, ASC-PG \citep{wang2017accelerating} and VRSC-PG~\citep{huo2018accelerated} (serving as a baseline for variance-reduced stochastic compositional optimization methods). 
We plot the objective function value gap and gradient norm against IFO complexity (measured by gradients calculation) for all four algorithms in two covariance settings and six real-world datasets. We observe that SARAH-Compositional outperforms all comparable algorithms.

The toy experiment provides evidence that our proposed SARAH-Compositional algorithm applied to risk-adverse portfolio management problem achieves state-of-the art performance.
Moreover, we note that due to the small mini-batch sizes, basic SCGD achieves a less satisfactory result, a phenomenon also shown by~\citet{huo2018accelerated,lian2017finite}.

\subsection{SARAH-Compositional Applied to Reinforcement Learning}\label{sec:exp_RL}
Next we demonstrate an experiment on reinforcement learning and test the performance of SARAH-Compositional on value function evaluation.
Let $V^\pi(s)$ be the value of state $s$ under policy $\pi$, then the value function $V^\pi(s)$ can be evaluated through Bellman equation:
\begin{eqnarray}
	\label{bellman}
	V^{\pi}(s_1) = \mathbb{E} [r_{s_1,s_2} + \gamma V^{\pi} (s_2)|s_1  ]
\end{eqnarray}
for all $ s_1,s_2, \dots, s_n\in \mathcal{S}$, where $\mathcal{S}$ represents the set of available states and $|\mathcal{S}| = n$.  In value function evaluation tasks, we minimize the square loss
\begin{equation}
\label{eq:rl}
 \sum_{s \in \mathcal{S}} \left( V^\pi(s) - \sum_{s' \in \mathcal{S}}P_{s, s'} \left( r_{s, s'} + \gamma V^\pi(s')\right)\right)^2
.
\end{equation}
We write $\sum_{s'\in \mathcal{S}} P_{s, s'}(r_{s, s'} + \gamma V^\pi(s'))$ as $\hat{V}^\pi(s)$. Equation~\eqref{eq:rl} is a special form of the stochastic compositional optimization problem by choosing $g(x)$ and $f(y)$ as follows~\citep{wang2017accelerating}:
\begin{align*}
g(s) &= \left[ V^\pi(s_1),\ldots, V^\pi(s_n), \hat{V}^\pi(s_1),\ldots, \hat{V}^\pi(s_n) \right],\\
f(s) &= \sum_{i = 1}^n (w_i - w_{n+i})^2,
\end{align*}
where $w\in \mathbb{R}^{2n}$ is the vector with the elements in $g(s)$ as components.

To model a reinforcement learning problem, we choose one of the commonly used setting of~\citet{dann2014policy} and generate a Markov decision process (MDP) with $400$ states and $10$ actions at each state. The transition probability is generated randomly from the uniform distribution $\mathcal{U}[0, 1]$ with $10^{-5}$ added to each component to ensure the ergodicity. In addition, the rewards $r_{s, s'}$ are sampled uniformly from $\mathcal{U}[0, 1]$.


\begin{figure*}
\hspace{-0.3in}
\centering
    \includegraphics[width=6.5in]{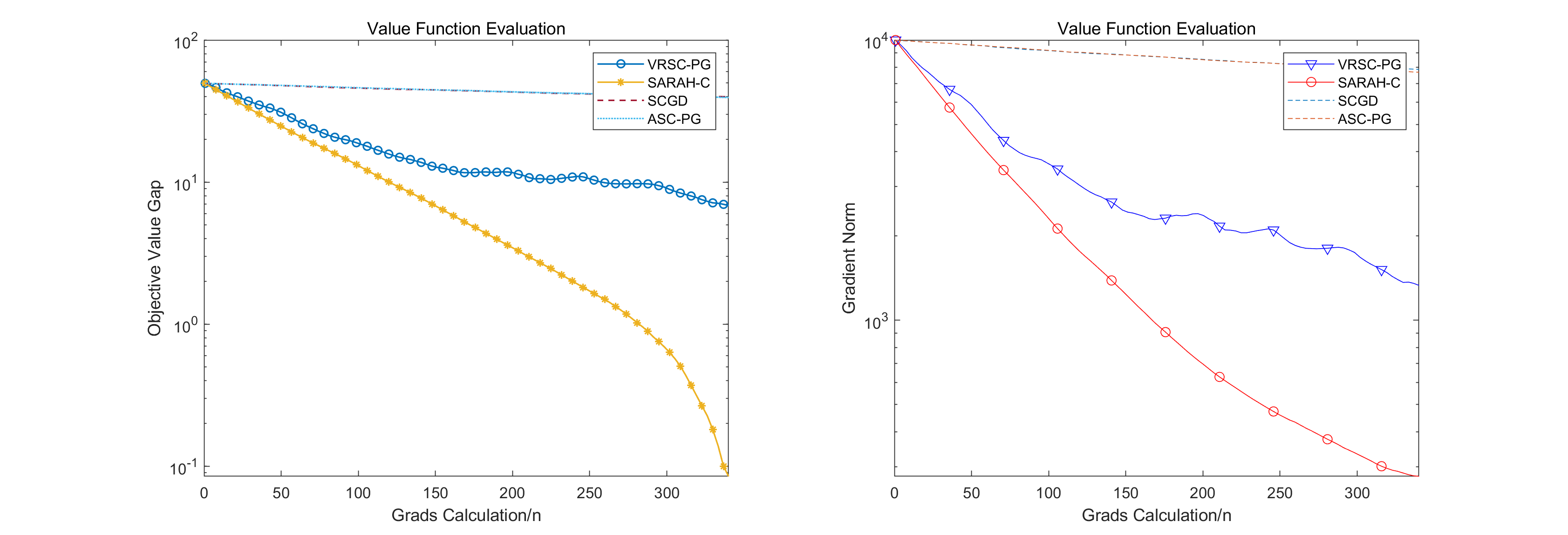}
  \caption{
  Experiment on the reinforcement learning. We plot the Objective Value Gap and Gradient Norm vs.~the IFO complexity (gradient calculation).
} 
\label{fig:rl}
\end{figure*}

We tested our results on different settings of batch size and inner iteration numbers.
In Figure~\ref{fig:rl} we plot our results on the batch size of $S^L_1 = S^L_2 = S^L_3 = 100$, $S_1 = S_2 = 5, S_3 = 1$, respectively.
The learning rate goes over the set $\{10^{-2}, 5\times 10^{-3}, 2\times 10^{-3}, 10^{-3}, 5\times 10^{-4}, 2\times 10^{-4}, 10^{-4}, 5\times 10^{-5}\},$ and the inner loop update iteration number $q$ are set to be 100.
We plot the objective value gap together with the gradient norm and use moving average to smooth the plot, which gives us Figure~\ref{fig:rl}.
From the figures we note that when the batch size is small and the iteration number is large, SARAH-Compositional outperforms VRSC on convergence speed, gradient norm and stability.
This supports our theoretical results and shows the advantage of SARAH-Compositional over VRSC on the effect of variance reduction.

\subsection{SARAH-Compositional Applied to SNE}\label{sec:tsne}
In SNE problem \citep{hinton2003stochastic} we use $z$'s to denote points in high dimensional space and $x$'s to denote their low dimensional images. We define
$$
p_{i | t}
=
\frac{\exp(- \|z_t - z_i\|^2 / 2 \sigma_i^2)}{\sum_{j = 1 , j \ne t}^n \exp(- \|z_t - z_j\|^2 / 2 \sigma_i^2)}
,\quad
q_{i | t}
=
\frac{\exp(- \|x_t - x_i\|^2)}{\sum_{j = 1, j \ne t}^n \exp(- \|x_t - x_j\|^2)}
,$$
where $\sigma_i$ controls the sensitivity to distance.
Then the SNE problem can be formulated as a non-convex compositional optimization problem \citep{liu2017variance} as \eqref{eq:problem} and \eqref{COOF}, where
$$
g_j(x)
=
\bigg[
x, e^{- \|x_1 - x_j\|^2} - 1, \dots, e^{- \|x_n - x_j\|^2} - 1
\bigg]^\top
,
$$
$$
f_i(w)
=
p_{i | 1} \left(
\|w_i - w_1\|^2 - \log(w_{n + 1})
\right)
+ \dots +
p_{i | n} \left(
\|w_i - w_n\|^2 - \log(w_{n + n})
\right)
.
$$
We implement SNE method on MNIST dataset, with sample size 2000 and dimension 784. We use SCGD\citep{wang2017stochastic}, ASC-PG\citep{wang2017accelerating}, and VRSC \citep{liu2017variance} as a baseline of variance reduced version of stochastic compositional optimization methods and compare its performance with SARAH-Compositional. We choose the best learning rate that keeps the algorithm to converge for each case.

\begin{figure}
\hspace{-0.3in}
\centering
    \includegraphics[width=6.5in]{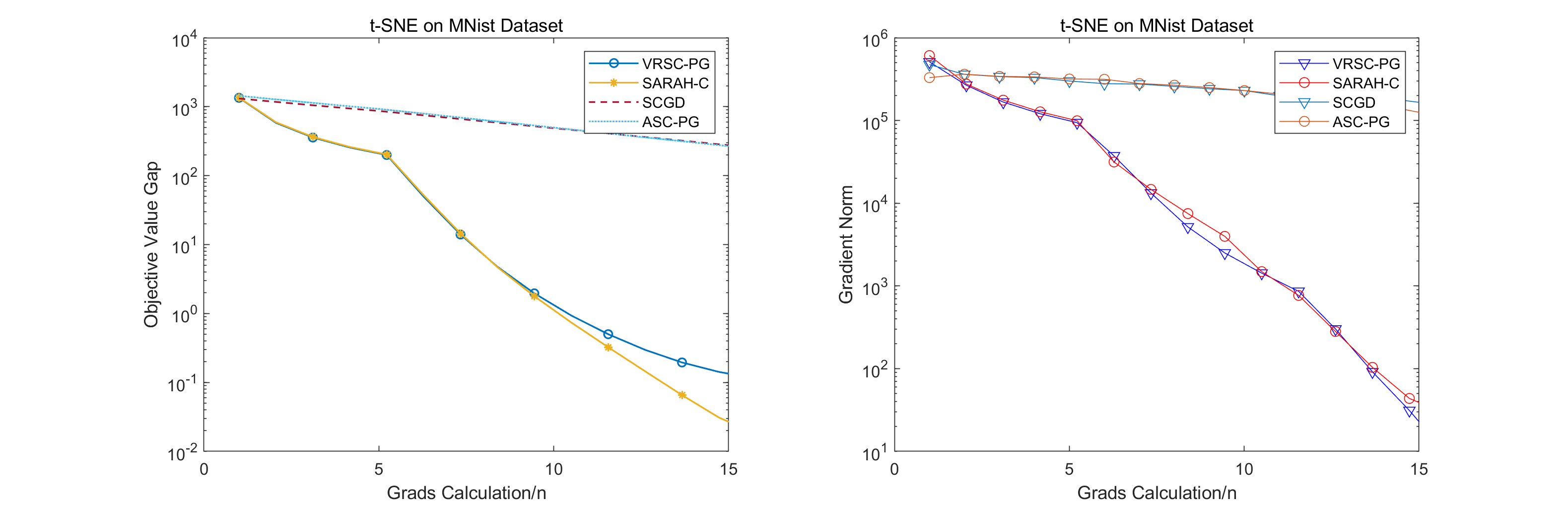}
    \caption{Experiment on SNE for MNIST dataset. The $x$-axis is the IFO complexity (gradient calculation) and the $y$-axis is the gradient norm.}
\label{fig:tsne}
\end{figure}

In our experiment, we choose a inner batch size of 5, an outer batch size of 1000, and optimal learning rate $10^{-5}$ for both algorithms.
In the left panel of Figure \ref{fig:tsne}, we plot the change of objective function value gap during iterations, and in the right panel we plot the gradient norm with respect to each outer loop update in SCGD, ASC-PG, VRSC and SARAH-Compositional. The left panel in Figure \ref{fig:tsne} shows that SARAH-Compositional has significantly better stability compared to VRSC. The gradient norm of SARAH-Compositional gradually decreases within each inner loop, while the gradient norm of VRSC accumulates within each inner loop and decreases at each outer loop.

We note that the objective function of t-SNE is non-convex. We observe from Figure \ref{fig:tsne} that, SARAH-Compositional outperforms VRSC with respect to the decrease in gradient norm against IFO complexity (gradient calculation), which is numerically consistent with our theory.

\section{Conclusion}\label{sec:conclusion}

In this paper, we propose a novel algorithm called SARAH-Compositional for solving stochastic compositional optimization problems using the idea of a recently proposed variance-reduced gradient method. 
Our algorithm achieves both outstanding theoretical and experimental results.
Theoretically, we show that the SARAH-Compositional algorithm can achieve desirable efficiency and IFO upper bound complexities for finding an $\varepsilon$-accurate solution of non-convex compositional problems in both finite-sum and online cases.
Experimentally, we compare our new compositional optimization method with a few rival algorithms for the task of portfolio management. 
Future directions include handling the non-smooth case and the theory of lower bounds for stochastic compositional optimization.
We hope this work can provide new perspectives to both optimization and machine learning communities interested in compositional optimization.

\bibliography{SmileSGD}

\begin{thebibliography}{33}
\providecommand{\natexlab}[1]{#1}
\providecommand{\url}[1]{\texttt{#1}}
\expandafter\ifx\csname urlstyle\endcsname\relax
  \providecommand{\doi}[1]{doi: #1}\else
  \providecommand{\doi}{doi: \begingroup \urlstyle{rm}\Url}\fi

\bibitem[Agarwal and Bottou(2015)]{agarwal2015lower}
Alekh Agarwal and Leon Bottou.
\newblock A lower bound for the optimization of finite sums.
\newblock In \emph{International Conference on Machine Learning}, pages 78--86,
  2015.

\bibitem[Allen-Zhu and Hazan(2016)]{allen2016variance}
Zeyuan Allen-Zhu and Elad Hazan.
\newblock Variance reduction for faster non-convex optimization.
\newblock In \emph{International Conference on Machine Learning}, pages
  699--707, 2016.

\bibitem[Arjevani et~al.(2019)Arjevani, Carmon, Duchi, Foster, Srebro, and
  Woodworth]{arjevani2019lower}
Yossi Arjevani, Yair Carmon, John~C Duchi, Dylan~J Foster, Nathan Srebro, and
  Blake Woodworth.
\newblock Lower bounds for non-convex stochastic optimization.
\newblock \emph{arXiv preprint arXiv:1912.02365}, 2019.

\bibitem[Dann et~al.(2014)Dann, Neumann, and Peters]{dann2014policy}
Christoph Dann, Gerhard Neumann, and Jan Peters.
\newblock Policy evaluation with temporal differences: A survey and comparison.
\newblock \emph{The Journal of Machine Learning Research}, 15\penalty0
  (1):\penalty0 809--883, 2014.

\bibitem[Defazio et~al.(2014)Defazio, Bach, and
  Lacoste-Julien]{defazio2014saga}
Aaron Defazio, Francis Bach, and Simon Lacoste-Julien.
\newblock Saga: A fast incremental gradient method with support for
  non-strongly convex composite objectives.
\newblock In \emph{Advances in neural information processing systems}, pages
  1646--1654, 2014.

\bibitem[Dentcheva et~al.(2017)Dentcheva, Penev, and
  Ruszczy{\'n}ski]{dentcheva2017statistical}
Darinka Dentcheva, Spiridon Penev, and Andrzej Ruszczy{\'n}ski.
\newblock Statistical estimation of composite risk functionals and risk
  optimization problems.
\newblock \emph{Annals of the Institute of Statistical Mathematics},
  69\penalty0 (4):\penalty0 737--760, 2017.

\bibitem[Fang et~al.(2018)Fang, Li, Lin, and Zhang]{fang2018spider}
Cong Fang, Chris~Junchi Li, Zhouchen Lin, and Tong Zhang.
\newblock Spider: Near-optimal non-convex optimization via stochastic
  path-integrated differential estimator.
\newblock In \emph{Advances in Neural Information Processing Systems}, pages
  689--699, 2018.

\bibitem[Ghadimi and Lan(2016)]{ghadimi2016accelerated}
Saeed Ghadimi and Guanghui Lan.
\newblock Accelerated gradient methods for nonconvex nonlinear and stochastic
  programming.
\newblock \emph{Mathematical Programming}, 156\penalty0 (1-2):\penalty0 59--99,
  2016.

\bibitem[Hinton and Roweis(2003)]{hinton2003stochastic}
Geoffrey~E Hinton and Sam~T Roweis.
\newblock Stochastic neighbor embedding.
\newblock In \emph{Advances in neural information processing systems}, pages
  857--864, 2003.

\bibitem[Huo et~al.(2018)Huo, Gu, Liu, and Huang]{huo2018accelerated}
Zhouyuan Huo, Bin Gu, Ji~Liu, and Heng Huang.
\newblock Accelerated method for stochastic composition optimization with
  nonsmooth regularization.
\newblock In \emph{Thirty-Second AAAI Conference on Artificial Intelligence},
  2018.

\bibitem[Lei et~al.(2017)Lei, Ju, Chen, and Jordan]{lei2017non}
Lihua Lei, Cheng Ju, Jianbo Chen, and Michael~I Jordan.
\newblock Non-convex finite-sum optimization via scsg methods.
\newblock In \emph{Advances in Neural Information Processing Systems}, pages
  2345--2355, 2017.

\bibitem[Li and Lin(2015)]{li2015accelerated}
Huan Li and Zhouchen Lin.
\newblock Accelerated proximal gradient methods for nonconvex programming.
\newblock In \emph{Advances in neural information processing systems}, pages
  379--387, 2015.

\bibitem[Lian et~al.(2017)Lian, Wang, and Liu]{lian2017finite}
Xiangru Lian, Mengdi Wang, and Ji~Liu.
\newblock Finite-sum composition optimization via variance reduced gradient
  descent.
\newblock In \emph{International Conference on Artificial Intelligence and
  Statistics}, pages 1159--1167, 2017.

\bibitem[Lin et~al.(2018)Lin, Fan, Wang, and Jordan]{lin2018improved}
Tianyi Lin, Chenyou Fan, Mengdi Wang, and Michael~I Jordan.
\newblock Improved oracle complexity for stochastic compositional variance
  reduced gradient.
\newblock \emph{arXiv preprint arXiv:1806.00458}, 2018.

\bibitem[Liu et~al.(2017)Liu, Liu, and Tao]{liu2017variance}
Liu Liu, Ji~Liu, and Dacheng Tao.
\newblock Variance reduced methods for non-convex composition optimization.
\newblock \emph{arXiv preprint arXiv:1711.04416}, 2017.

\bibitem[Nesterov(2004)]{NESTEROV}
Yurii Nesterov.
\newblock \emph{Introductory lectures on convex optimization: A basic course},
  volume~87.
\newblock Springer, 2004.

\bibitem[Nguyen et~al.(2017)Nguyen, Liu, Scheinberg, and
  Tak{\'a}{\v{c}}]{nguyen2017sarah}
Lam~M Nguyen, Jie Liu, Katya Scheinberg, and Martin Tak{\'a}{\v{c}}.
\newblock Sarah: A novel method for machine learning problems using stochastic
  recursive gradient.
\newblock In \emph{International Conference on Machine Learning}, pages
  2613--2621, 2017.

\bibitem[Nguyen et~al.(2019)Nguyen, van Dijk, Phan, Nguyen, Weng, and
  Kalagnanam]{nguyen2019optimal}
Lam~M Nguyen, Marten van Dijk, Dzung~T Phan, Phuong~Ha Nguyen, Tsui-Wei Weng,
  and Jayant~R Kalagnanam.
\newblock Optimal finite-sum smooth non-convex optimization with sarah.
\newblock \emph{arXiv preprint arXiv:1901.07648}, 2019.

\bibitem[Reddi et~al.(2016)Reddi, Hefny, Sra, Poczos, and
  Smola]{reddi2016stochastic}
Sashank~J Reddi, Ahmed Hefny, Suvrit Sra, Barnabas Poczos, and Alex Smola.
\newblock Stochastic variance reduction for nonconvex optimization.
\newblock In \emph{International conference on machine learning}, pages
  314--323, 2016.

\bibitem[Schmidt et~al.(2017)Schmidt, Le~Roux, and Bach]{schmidt2017minimizing}
Mark Schmidt, Nicolas Le~Roux, and Francis Bach.
\newblock Minimizing finite sums with the stochastic average gradient.
\newblock \emph{Mathematical Programming}, 162\penalty0 (1-2):\penalty0
  83--112, 2017.

\bibitem[Shalev-Shwartz and Zhang(2013)]{shalev2013stochastic}
Shai Shalev-Shwartz and Tong Zhang.
\newblock Stochastic dual coordinate ascent methods for regularized loss
  minimization.
\newblock \emph{Journal of Machine Learning Research}, 14\penalty0
  (Feb):\penalty0 567--599, 2013.

\bibitem[Shalev-Shwartz and Zhang(2014)]{shalev2014accelerated}
Shai Shalev-Shwartz and Tong Zhang.
\newblock Accelerated proximal stochastic dual coordinate ascent for
  regularized loss minimization.
\newblock In \emph{International Conference on Machine Learning}, pages 64--72,
  2014.

\bibitem[Shapiro et~al.(2009)Shapiro, Dentcheva, and
  Ruszczy{\'n}ski]{SHAPIRO-DENTCHEVA-RUSZCZYNSKI}
Alexander Shapiro, Darinka Dentcheva, and Andrzej Ruszczy{\'n}ski.
\newblock \emph{Lectures on stochastic programming: modeling and theory}.
\newblock SIAM, 2009.

\bibitem[Sutton and Barto(1998)]{SUTTON-BARTO}
Richard~S Sutton and Andrew~G Barto.
\newblock \emph{Reinforcement learning: An introduction}.
\newblock 1998.

\bibitem[Wang et~al.(2017{\natexlab{a}})Wang, Fang, and
  Liu]{wang2017stochastic}
Mengdi Wang, Ethan~X Fang, and Han Liu.
\newblock Stochastic compositional gradient descent: Algorithms for minimizing
  compositions of expected-value functions.
\newblock \emph{Mathematical Programming}, 161\penalty0 (1-2):\penalty0
  419--449, 2017{\natexlab{a}}.

\bibitem[Wang et~al.(2017{\natexlab{b}})Wang, Liu, and
  Fang]{wang2017accelerating}
Mengdi Wang, Ji~Liu, and Ethan~X Fang.
\newblock Accelerating stochastic composition optimization.
\newblock \emph{Journal of Machine Learning Research}, 18:\penalty0 1--23,
  2017{\natexlab{b}}.

\bibitem[Wang et~al.(2019)Wang, Ji, Zhou, Liang, and
  Tarokh]{wang2019spiderboost}
Zhe Wang, Kaiyi Ji, Yi~Zhou, Yingbin Liang, and Vahid Tarokh.
\newblock Spiderboost and momentum: Faster variance reduction algorithms.
\newblock In \emph{Advances in Neural Information Processing Systems}, pages
  2403--2413, 2019.

\bibitem[Woodworth and Srebro(2016)]{woodworth2016tight}
Blake~E Woodworth and Nati Srebro.
\newblock Tight complexity bounds for optimizing composite objectives.
\newblock In \emph{Advances in Neural Information Processing Systems}, pages
  3639--3647, 2016.

\bibitem[Xiao and Zhang(2014)]{xiao2014proximal}
Lin Xiao and Tong Zhang.
\newblock A proximal stochastic gradient method with progressive variance
  reduction.
\newblock \emph{SIAM Journal on Optimization}, 24\penalty0 (4):\penalty0
  2057--2075, 2014.

\bibitem[Yang et~al.(2019)Yang, Wang, and Fang]{yang2019multilevel}
Shuoguang Yang, Mengdi Wang, and Ethan~X Fang.
\newblock Multilevel stochastic gradient methods for nested composition
  optimization.
\newblock \emph{SIAM Journal on Optimization}, 29\penalty0 (1):\penalty0
  616--659, 2019.

\bibitem[Zhang and Xiao(2019{\natexlab{a}})]{zhang2019multi}
Junyu Zhang and Lin Xiao.
\newblock Multi-level composite stochastic optimization via nested variance
  reduction.
\newblock \emph{arXiv preprint arXiv:1908.11468}, 2019{\natexlab{a}}.

\bibitem[Zhang and Xiao(2019{\natexlab{b}})]{zhang2019stochastic}
Junyu Zhang and Lin Xiao.
\newblock A stochastic composite gradient method with incremental variance
  reduction.
\newblock \emph{arXiv preprint arXiv:1906.10186}, 2019{\natexlab{b}}.

\bibitem[Zhou et~al.(2018)Zhou, Xu, and Gu]{zhou2018stochastic}
Dongruo Zhou, Pan Xu, and Quanquan Gu.
\newblock Stochastic nested variance reduced gradient descent for nonconvex
  optimization.
\newblock In \emph{Advances in Neural Information Processing Systems}, pages
  3921--3932, 2018.

\end{thebibliography}

\newpage

\section{Detailed Analysis of Convergence Theorems}\label{sec:app_proof}
In this section, we detail the analysis of our Theormems \ref{theo:SFOSPfinitesum} and \ref{theo:SFOSPonline}. Before moving on, we first provide a key lemma that serves as their common analysis, whose proof is provided in \S\ref{sec:proof,lemm:boundedgrad2}. 
We assume that the expected estimation error squared is bounded as the following for any $t$ and some parameters $\omega_1$, $\omega_2$ and $\omega_3$ to be specified here:

\begin{lemma}\label{lemm:boundedgrad2}
Assume that for any initial point $x_0\in \RR^d$
\beq\label{omegas}
\begin{aligned}
    &
\EE\| \bF_0 - (\bG_0)^{\top} \nabla f(\bg_0) \|^2
=
\EE\left\|
   \left(\bG_0\right)^{\top} \left[ \dfrac{1}{S_3}  \sum\limits_{i\in \mathcal{S}^3_{0}} \nabla f_{i}(\bg_0) \right]   - (\bG_0)^{\top} \nabla f(\bg_0) 
\right\|^2
\le \omega_1
  ,  \\&
\EE\| \bG_0 - \partial g (x_0) \|^2
=
\EE \left\|
\frac{1}{S_2} \sum\limits_{j\in \mathcal{S}^2_{0}} \partial g_j(x_0)
-
\partial g (x_0) 
\right\|^2
   \le 
\omega_2
  ,  \\&
\EE\| \bg_0 - g (x_0) \|^2
  =
\EE\left\| 
\frac{1}{S_1} \sum\limits_{j\in \mathcal{S}^1_{0}} g_j(x_0) - g (x_0) \right\|^2
   \le  
\omega_3
   .
\end{aligned}\eeq
Then we have
\beq \label{boundedgrad2}
\begin{aligned}
  &
\EE \|\nabla \Phi(\widetilde{x})\|^2
 \le
    \frac{2 }{T\eta} [ \Phi(x_0) - \Phi^\ast ]
+
     3(\omega_1  +  M_f^2 \omega_2  +  M_g^2 L_f^2 \omega_3)
      .
\end{aligned}
\eeq
\end{lemma}

With Lemma \ref{lemm:boundedgrad2}, our goal is to make the left hand of \eqref{boundedgrad2} no greater than $\cO(\varepsilon^2)$.
We present the proofs for finite-sum and online cases, separately.

\subsection{Proof of Theorem~\ref{theo:SFOSPfinitesum}}
\begin{proof}[Proof of Theorem~\ref{theo:SFOSPfinitesum}]
In this finite-sum case for Algorithm \ref{Alg:SARAH-SCGD}, $\omega_1 = \omega_2 = \omega_3 = 0$.
Bringing into \eqref{boundedgrad2}, in order to achieve $\leq \varepsilon^2$ for $ \mathbb{E}[ \| \nabla \Phi(\widetilde{x})\|^2 ]$ we need $\dfrac{2}{T\eta}\left( \Phi(x_{0})  - \Phi^\ast \right) \le \varepsilon^2$.
Recalling the choice of stepsize in \eqref{etace}, the total iteration complexity $Q_{\text{iter}}$ is
\beq\label{Tdef}
\begin{aligned}
Q_{\text{iter}}
     =
     \frac{2[ \Phi(x_0) - \Phi^\ast] }{\varepsilon^2} \cdot \eta^{-1}
     =
     \frac{2[ \Phi(x_0) - \Phi^\ast] }{\varepsilon^2} \cdot
     \max\left( \sqrt{6S_o q},  2L_\Phi \right)
,
\end{aligned}\eeq
proving \eqref{Qiter}.
When \eqref{bscond} holds i.e.~$\sqrt{6S_o q} \ge 2L_\Phi$ the first term in the max of \eqref{Tdef} dominates.
Note that $S^L_1=m=S^L_2, S^L_3 = n$, the total IFO complexity achieving $\varepsilon$-accurate solution is hence bounded by
\begin{align*}
\text{IFO complexity}
    & =
(S^L_1+S^L_2+S^L_3)  \left\lceil   \frac{Q_{\text{iter}}}{q} \right\rceil
+
2(S_1+S_2+S_3) \left(
 Q_{\text{iter}} - \left\lceil   \frac{Q_{\text{iter}}}{q} \right\rceil
\right)
    \\&\le
S^L_1+S^L_2+S^L_3
 + 
\left(   \frac{S^L_1+S^L_2+S^L_3}{q}  +  2(S_1+S_2+S_3)  \right) Q_{\text{iter}} 
    \\& \le
2m+n
 + 
\left(   \frac{2m+n}{q}  +  2(S_1+S_2+S_3)  \right) \cdot \frac{2[ \Phi(x_0) - \Phi^\ast] }{\varepsilon^2} \cdot \sqrt{6S_o q}
    \\& \le
2m+n
 + 
\sqrt{2m + n}
 \cdot
\sqrt{
   (S_1+S_2+S_3 )   
  \left(1 + \frac{2}{S_3}\right) \left(\frac{M_g^4L_f^2}{S_1} + \frac{M_f^2 L_g^2}{S_2}\right)
}
\\&
\cdot
  \frac{\sqrt{216} [ \Phi(x_0) - \Phi^\ast] }{\varepsilon^2}
      ,
\end{align*}
where in the last step we plugged in $S_o$ and $q$ as in \eqref{Soqeq}.
This completes our proof of \eqref{IFObound} and the whole theorem.

\end{proof}

\subsection{Proof of Theorem~\ref{theo:SFOSPonline}}
\begin{proof}[Proof of Theorem~\ref{theo:SFOSPonline}]
In the online case for Algorithm~\ref{Alg:SARAH-SCGD} we need both $\frac{2}{T\eta}(\Phi(x_0) - \Phi^*) \leq \varepsilon^2$ and $3(\omega_1 + M_f^2\omega_2 + M_g^2 L_f^2 \omega_3) \leq \varepsilon^2$ to achieve $\EE[\|\nabla \Phi(\tilde{x})\|^2] \leq 2 \varepsilon^2$ (factor 2 here is for notation consistency) Recalling~\eqref{etace2}, the total iteration complexity $Q_{\text{iter}}$ is 
\beq\label{Tdef2}
\begin{aligned}
Q_{\text{iter}}
     =
     \frac{2[ \Phi(x_0) - \Phi^\ast] }{\varepsilon^2} \cdot \eta^{-1}
     =
     \frac{2[ \Phi(x_0) - \Phi^\ast] }{\varepsilon^2} \cdot
     \max\left( \sqrt{6S_o q},  2L_\Phi \right)
,
\end{aligned}\eeq
proving~\eqref{Qiter2}. When~\eqref{bscond2} holds i.e.~$\sqrt{6S_0 q} \geq 2L_\Phi$ the first term in the max of~\eqref{Tdef2} dominates. Note that $S^L_1 = \frac{3H_3 M_g^2L_f^2}{\varepsilon^2}, S^L_2 = \frac{3H_2 M_f^2}{\varepsilon^2}, S^L_3 = \frac{3H_1 M_g^2}{ \varepsilon^2}$, the total IFO complexity achieving $\varepsilon$-approximate stationary point is hence bounded by
\begin{align*}
\text{IFO complexity}
    & =
(S^L_1+S^L_2+S^L_3)  \left\lceil   \frac{Q_{\text{iter}}}{q} \right\rceil
+
2(S_1+S_2+S_3) \left(
 Q_{\text{iter}} - \left\lceil   \frac{Q_{\text{iter}}}{q} \right\rceil
\right)
    \\&\le
S^L_1+S^L_2+S^L_3
 + 
\left(   \frac{S^L_1+S^L_2+S^L_3}{q}  +  2(S_1+S_2+S_3)  \right) Q_{\text{iter}} 
    \\& \le
\frac{3H_3M_g^2L_f^2 + 3H_2M_f^2 + 3H_1M_g^2}{\varepsilon^2}
\\ & + 
\left(   \frac{3H_3M_g^2L_f^2 + 3H_2M_f^2 + 3H_1M_g^2}{\varepsilon^2q}+  2(S_1+S_2+S_3)  \right) \cdot \frac{2[ \Phi(x_0) - \Phi^\ast] }{\varepsilon^2} \cdot \sqrt{6S_o q}
    \\& \le
\frac{D_0}{\varepsilon^2}
 + 
\sqrt{D_0} 
 \cdot 
\sqrt{(S_1 + S_2 + S_3)\left(1 + \dfrac{2}{S_3} \right)  \left(\dfrac{M_g^4L_f^2}{S_1} + \dfrac{M_f^2 L_g^2}{S_2}\right)}\cdot \frac{\sqrt{216}[ \Phi(x_0) - \Phi^\ast] }{\varepsilon^3}
      ,
\end{align*}
where in the last step we plugged in $S_o$, $q$ as in \eqref{Soqeq2} and $D_0$ as in \eqref{D_0}.
This completes our proof of \eqref{IFObound2} and the whole theorem.

\end{proof}

\subsection{Proof of Lemma~\ref{lemm:boundedgrad2}}\label{sec:proof,lemm:boundedgrad2}

Before starting our proof, we first prove the following three lemmas that characterize the error bounds induced by our iterations.
For approximation errors for $\bg$-iteration and $\bG$-iteration, we prove the following
\begin{lemma}[Error bound induced by $\bg$-iteration]\label{lemm:approxg}
We have for any fixed $t\ge 0$
\beq\label{approxg}
\EE \left\|
\nabla \Phi\left(x_t\right)
-
\left(\partial g\left(x_t\right)\right)^{\top} \nabla f\left(\bg_t\right)
\right\|^2
    \le
    M_g^2 L_f^2 \omega_3 +
    \frac{M_g^4 L_f^2}{S_1}\cdot \eta^2 \sum_{s=\lfloor t/q \rfloor q + 1}^t \EE\| \bF_{s-1} \|^2
    .
\eeq
where the expectation in the last term is taken over a uniformly chosen $j$ in $[1,m]$.
\end{lemma}

In addition, we have
\begin{lemma}[Error bound induced by $\bG$-iteration]\label{lemm:Gapprox}
\label{lemm:Gapprox}
We have for any fixed $t \ge 0$
\beq\label{Gapprox}
\EE\left\|
\left(\partial g(x_t)\right)^\top \nabla f(\bg_t)
-
\left(\bG_t\right)^\top \nabla f(\bg_t)
\right\|^2
 \le
M_f^2 \omega_2 + \frac{M_f^2 L_g^2}{S_2}\cdot \eta^2 \sum_{s=\lfloor t/q \rfloor q + 1}^t\EE\|\bF_{s-1}\|^2
.
\eeq
where the expectation in the last term is taken over a uniformly chosen $j$ in $[1,m]$.
\end{lemma}

Finally, we prove the following lemma that characterizes the gap between $\bF$-iteration and $\left(\bG_t\right)^{\top} \nabla f\left(\bg_t\right)$:
\begin{lemma}[Error bound induced by $\bF$-iteration]\label{lemm:Fapprox}
We have for any fixed $t \ge 0$
\beq\label{Fapprox}
\EE\left\|
\left(\bG_t\right)^{\top} \nabla f\left(\bg_t\right)
-
\bF_t
\right\|^2
 \le
\omega_1
+
\frac{2}{S_3} \left(\frac{M_g^4L_f^2}{S_1} + \frac{M_f^2 L_g^2}{S_2}\right)\cdot \eta^2 \sum_{s=\lfloor t/q \rfloor q + 1}^t \mathbb{E}\left\|\bF_{s-1}\right\|^2
,
\eeq
where the expectation in the last term is taken over a uniformly chosen $i$ in $[1,n]$.
\end{lemma}

We prove Lemma~\ref{lemm:boundedgrad2} in the following steps
\begin{enumerate}[(i)]
\item
Standard arguments along with the smoothness Assumption \ref{Assumption:Smoothness}, we have from the update rule is $x_{t+1} = x_t - \eta \bF_t$ and \eqref{assmoothness} that for any $x,x'\in \RR^d$,

$$\begin{aligned}
\left\| \nabla \Phi(x) - \nabla \Phi(x') \right\|
 &=
\left\| [\partial g(x)]^\top\nabla f(g(x)) - [\partial g(x')]^\top\nabla f(g(x'))\right\|
 \\&\le
\frac{1}{nm}\sum_{j=1}^m \sum_{i=1}^n 
\left\|[\partial g_j(x)]^\top\nabla f_i(g(x)) - [\partial g_j(x')]^\top\nabla f_i(g(x'))\right\|
 \\&\le 
L_\Phi \|x-x'\|
,
\end{aligned}$$
and hence a Taylor's expansion argument gives

\begin{align*}
\Phi(x_{t+1})
&\le
\Phi(x_t) + (\nabla \Phi(x_t))^\top (x_{t+1} - x_t) + \frac{L_\Phi}{2} \|x_{t+1} - x_t\|^2
\\&=
\Phi(x_t) - \frac{\eta}{2} \cdot 2 (\nabla \Phi(x_t))^\top \bF_t + \frac{L_\Phi \eta^2}{2} \|\bF_t\|^2
\\&\le 
\Phi(x_t) - \frac{\eta}{2} \| \nabla \Phi(x_t)\|^2
+ \frac{\eta}{2} \| \nabla \Phi(x_t) - \bF_t \|^2
- \left( \frac{\eta}{2} - \frac{L_\Phi\eta^2}{2} \right) \| \bF_t \|^2, \label{eq_proof_lem_000}
\\&\le 
\Phi(x_t) - \frac{\eta}{2} \| \nabla \Phi(x_t)\|^2
+ \frac{\eta}{2} \left(
\| \nabla \Phi(x_t) - \bF_t \|^2 - \frac12 \| \bF_t \|^2
\right)
,
\end{align*}
where the second to last inequality above follows from \eqref{etace} and the fact $2a^Tb = \|a\|^2 - \|a-b\|^2 + \|b\|^2$ for any real vectors $a,b$, and the last inequality is due to $L_\Phi \eta \le  1/2$ given by \eqref{etace2}. 
Summing the above over $t = 0,\ldots,T-1$ and taking expectation on both sides allow us to conclude
\beq
\begin{aligned}
\Phi^\ast \le  \mathbb{E}[ \Phi(x_T) ]
 &\leq
\Phi(x_0) - \frac{\eta}{2} \sum_{t=0}^{T-1} \mathbb{E}[ \| \nabla \Phi(x_{t})\|^2 ]
\\ & \qquad
+ \frac{\eta}{2} \left( \sum_{t=0}^{T-1} \mathbb{E}[ \| \bF_{t} - \nabla \Phi(x_{t}) \|^2 ]
- \frac12 \sum_{t=0}^{T-1} \mathbb{E} [ \| \bF_{t} \|^2 ] \right)
.
\end{aligned}
 \label{eq_proof_lem_000}
\eeq
Since $\widetilde{x}$ is chosen uniformly at random from $\{x_t\}_{t=0}^{T-1}$, rearranging \eqref{eq_proof_lem_000} gives
\beq \label{boundedgrad}
\begin{aligned}
\EE \|\nabla \Phi(\widetilde{x})\|^2 
&= 
\frac{1}{T} \sum_{t=0}^{T-1}  \EE[ \| \nabla \Phi(x_{t})\|^2 ]
   \\ &\leq
    \frac{2 }{T\eta} [ \Phi(x_0) - \Phi^\ast ]
       + \frac{1}{T}\left( \sum_{t=0}^{T-1} \mathbb{E}[ \| \bF_{t} - \nabla \Phi(x_{t}) \|^2 ]
       - \frac12 \sum_{t=0}^{T-1} \mathbb{E} [ \| \bF_{t} \|^2 ] \right)
       .
\end{aligned}
\eeq

\item
To bound $\|\bF_t - \nabla \Phi(x_t) \|^2$ in expectation, note
\begin{equation}\label{eq:7}
\begin{aligned}
 &\quad
\mathbb{E}\left\|\bF_{t}-\nabla \Phi\left(x_{t}\right)\right\|^{2}
  \\&=
\mathbb{E}\left\|
\bF_{t}-\left(\bG_t\right)^{\top}\nabla f(\bg_t)
+ \left(\bG_t\right)^{\top}\nabla f(\bg_t) -\left(\partial g\left(x_{t}\right)\right)^{\top} \nabla f\left(\bg_{t}\right)
+\left(\partial g\left(x_{t}\right)\right)^{\top} \nabla f\left(\bg_{t}\right)-\nabla \Phi\left(x_{t}\right)
\right\|^{2}
  \\ & \le
3 \mathbb{E}\left\|\bF_{t}-\left(\bG_t\right)^{\top} \nabla f\left(\bg_{t}\right)\right\|^{2}
+ 
3 \EE \left\|\left(\bG_t\right)^{\top}\nabla f(\bg_t) - \left(\partial g\left(x_t\right)\right)^{\top}\nabla f(\bg_t)\right\|^2
 \\ &\quad
+
3 \mathbb{E}\left\|\left(\partial g\left(x_t\right)\right)^{\top} \nabla f\left(\bg_{t}\right)-\nabla \Phi\left(x_{t}\right)\right\|^{2}
,
\end{aligned}
\end{equation}
where in the last inequality we applied Minkowski's inequality (along with elementary algebra). 
The three terms in~\eqref{eq:7} can be estimated using a combination of Lemmas \ref{lemm:approxg}, \ref{lemm:Gapprox} and \ref{lemm:Fapprox}, where we have

\begin{equation}
\mathbb{E}\left\|\bF_{t}-\nabla \Phi(x_{t})\right\|^{2}
 \le
3\left(\omega_1 + M_f^2 \omega_2 + M_g^2 L_f^2 \omega_3\right)
+
3S_o \eta^2\sum_{s=\lfloor t/q \rfloor q + 1}^t \mathbb{E}\left\|\bF_{s-1}\right\|^{2}
,
\label{summed}
\end{equation}
where $S_o =\left(1 + \dfrac{2}{S_3} \right)  \left(\dfrac{M_g^4L_f^2}{S_1} + \dfrac{M_f^2 L_g^2}{S_2}\right)$ defined in \eqref{Soqeq}.
Back to \eqref{boundedgrad} which is repeated in below
\beq \tag{\eqref{boundedgrad}}
\begin{aligned}
\EE \|\nabla \Phi(\widetilde{x})\|^2 
&= 
\frac{1}{T} \sum_{t=0}^{T-1}  \EE[ \| \nabla \Phi(x_{t})\|^2 ]
   \\ &\leq
    \frac{2 }{T\eta} [ \Phi(x_0) - \Phi^\ast ]
       + \frac{1}{T}\left( \sum_{t=0}^{T-1} \mathbb{E}[ \| \bF_{t} - \nabla \Phi(x_{t}) \|^2 ]
       - \frac12 \sum_{t=0}^{T-1} \mathbb{E} [ \| \bF_{t} \|^2 ] \right)
       .
\end{aligned}
\eeq
For the second term of \eqref{boundedgrad}, note that aggregating \eqref{summed} for each $q$-step epoch implies, when stepsize $\eta$ picked as in \eqref{etace} satisfies $3S_o q \eta^2 - 1/2 \le 0$, the following
\beq\label{mical2}  \begin{aligned}
&\quad
     \frac{1}{T}\left(\sum_{t=0}^{T-1} \mathbb{E}[ \| \bF_{t} - \nabla \Phi(x_{t}) \|^2 ] 
     - 
     \frac12 \sum_{t=0}^{T-1} \mathbb{E} [ \| \bF_{t} \|^2 ]\right)
\\&  \le
     \frac{1}{T} \sum_{t=0}^{T-1} \left(
     3\left(\omega_1 + M_f^2 \omega_2 + M_g^2 L_f^2 \omega_3\right)
+
3S_o \eta^2\sum_{s=\lfloor t/q \rfloor q + 1}^t \mathbb{E}\left\|\bF_{s-1}\right\|^{2}
     \right)
     - 
     \frac12  \cdot \frac{1}{T}\sum_{t=0}^{T-1} \mathbb{E} [ \| \bF_{t} \|^2 ]
\\&    \overset{\eqref{summed}}{\le}
     \frac{1}{T}\cdot T \cdot 3(\omega_1  +  M_f^2 \omega_2  +  M_g^2 L_f^2 \omega_3)
    + 3S_o q\eta^2\cdot \frac{1}{T}  \sum_{t =0}^{T-1} \mathbb{E} \| \bF_{t} \|^2
    - \frac{1}{2}\cdot \frac{1}{T} \sum_{t=0}^{T-1} \mathbb{E} \| \bF_{t} \|^2
\\& =
     3(\omega_1  +  M_f^2 \omega_2  +  M_g^2 L_f^2 \omega_3)
         + \Big[ 3S_oq\eta^2 -  \frac12 \Big] \cdot\frac{1}{T} \sum_{t=0}^{T-1} \mathbb{E}\| \bF_{t} \|^2
\\& \le
     3(\omega_1  +  M_f^2 \omega_2  +  M_g^2 L_f^2 \omega_3)
        .
  \end{aligned}
  \eeq
Finally, \eqref{boundedgrad} and \eqref{mical2} together conclude~\eqref{boundedgrad2} and hence Lemma~\ref{lemm:boundedgrad2}.
\end{enumerate}


\subsection{Proof of Corollary \ref{coro:OPTBSfinitesum}}
\begin{proof}[Proof of Corollary \ref{coro:OPTBSfinitesum}]
How to optimize the batch size?
First, we minimize the IFO complexity with respect to $S_3$.
This is equivalent to minimize $(S_1+S_2+S_3 )     \left(1 + \frac{2}{S_3}\right)$, which is simply $S_3 = \sqrt{2(S_1+S_2 )}$.
The problem further reduces to the following
$$\begin{aligned}
&\quad
\text{minimize}_{S_1,S_2,S_3}  (S_1+S_2+S_3 )
  \left(1 + \frac{2}{S_3}\right) \left(\frac{M_g^4L_f^2}{S_1} + \frac{M_f^2 L_g^2}{S_2}\right)
\\&=
\text{minimize}_{S_1,S_2} \left.
   (S_1+S_2+S_3 )   
  \left(1 + \frac{2}{S_3}\right) \left(\frac{M_g^4L_f^2}{S_1} + \frac{M_f^2 L_g^2}{S_2}\right)
 \right|_{S_3 = \sqrt{2(S_1+S_2 )}}
\\&=
\text{minimize}_{S_1,S_2}
 \left(\sqrt{S_1+S_2}+\sqrt{2} \right)^2 \left(\frac{M_g^4L_f^2}{S_1} + \frac{M_f^2 L_g^2}{S_2}\right)
 .
\end{aligned}$$
Taking partial derivatives w.r.t~$S_1$ and $S_2$ in above respectively we have
$$\begin{aligned}
&\quad
2 \left(\sqrt{S_1+S_2}+\sqrt{2} \right) \frac{1}{2\sqrt{S_1+S_2}}
 \left(\frac{M_g^4L_f^2}{S_1} + \frac{M_f^2 L_g^2}{S_2}\right)
+
 \left(\sqrt{S_1+S_2}+\sqrt{2} \right)^2 
 \left(-\frac{M_g^4L_f^2}{S_1^2}\right)
\\
& =0=
2 \left(\sqrt{S_1+S_2}+\sqrt{2} \right) \frac{1}{2\sqrt{S_1+S_2}}
 \left(\frac{M_g^4L_f^2}{S_1} + \frac{M_f^2 L_g^2}{S_2}\right)
+
 \left(\sqrt{S_1+S_2}+\sqrt{2} \right)^2 
 \left(-\frac{M_f^2 L_g^2}{S_2^2}\right)
\end{aligned}$$
so
$$
\frac{M_g^4L_f^2}{S_1^2}
 =
\frac{M_f^2 L_g^2}{S_2^2}
  \quad\Longrightarrow\quad
  \frac{S_1}{ M_g^2 L_f} 
=
  \frac{S_2}{M_f L_g}
=
  \frac{S_3^2}{2(M_g^2 L_f + M_f L_g)}
\equiv
 Z
,
$$
which further reduces to
$$\begin{aligned}
 \left( \sqrt{S_1+S_2}+\sqrt{2} \right)^2 \left(\frac{M_g^4L_f^2}{S_1} + \frac{M_f^2 L_g^2}{S_2} \right)
  &=
 \left( \sqrt{Z\cdot M_g^2 L_f+Z\cdot M_f L_g}+\sqrt{2} \right)^2
\\&\quad\cdot
 \left(\frac{M_g^4L_f^2}{Z\cdot M_g^2 L_f} + \frac{M_f^2 L_g^2}{Z\cdot M_f L_g} \right)
  \\&=
 \left( \sqrt{ M_g^2 L_f +  M_f L_g}+\sqrt{\frac{2}{Z}} \right)^2 \left(M_g^2 L_f + M_f L_g\right)
  \\&=
\left( \sqrt{ L_\Phi }+\sqrt{\frac{2}{Z}} \right)^2 L_\Phi
 =
\frac{\left(  L_\Phi Z +\sqrt{2 L_\Phi Z}\right)^2 }{ Z^2}
.
\end{aligned}$$
Therefore we aim to minimize the above quantity with respect to $Z$ and further IFO.
It is easy to verify (and we omit the details) that at maximal, $Z = \sqrt{\frac{3(2m+n)}{2L_\Phi^2}}$ for the finite-sum case and $Z = \sqrt{\frac{3D_0}{2L_\Phi^2 \varepsilon^2}}$, so
$$\begin{aligned}
 \left( \sqrt{S_1+S_2}+\sqrt{2} \right)^2 \left(\frac{M_g^4L_f^2}{S_1} + \frac{M_f^2 L_g^2}{S_2} \right)
  &=
\left( \sqrt{ L_\Phi }+\sqrt{\frac{2}{Z}} \right)^2 L_\Phi
 =
\left( \sqrt{ L_\Phi }+\sqrt[4]{\frac{8L_\Phi^2}{3(2m+n)}} \right)^2 L_\Phi
 \\&=
\left( 1 + \sqrt[4]{\frac{8}{3(2m+n)}}  \right)^2 L_\Phi^2
,
\end{aligned}$$
in the finite-sum case and
$$
 \left( \sqrt{S_1+S_2}+\sqrt{2} \right)^2 \left(\frac{M_g^4L_f^2}{S_1} + \frac{M_f^2 L_g^2}{S_2} \right)
=
\left( 1 + \sqrt[4]{\frac{8\varepsilon^2}{3D_0}}  \right)^2 L_\Phi^2
,$$
in the online case.

So the IFO complexity in \eqref{IFObound} is
$$
2m+n
 + 
\sqrt{2m + n}
 \cdot
\left( 1 + \sqrt[4]{\frac{8}{3(2m+n)}}  \right) L_\Phi
\cdot
  \frac{\sqrt{216} [ \Phi(x_0) - \Phi^\ast] }{\varepsilon^2}
,
$$
proving \eqref{IFObound_optimal}.
In this case, the final choice of mini-batch sizes are
$$
  \frac{S_1}{ M_g^2 L_f} 
=
  \frac{S_2}{M_f L_g}
=
  \frac{S_3^2}{2 (M_f^2 L_g^2 + M_g^4 L_f^2)}
=
  \sqrt{\frac{3(2m+n)}{2L_\Phi^2}}
  ,
$$
leading to~\eqref{bcsize}, as long as these values are the minimizers in the feasible range.

Similarily in the online case, when $S_1, S_2, S_3$ are chosen to satisfy
$$
  \frac{S_1}{ M_g^2 L_f} 
=
  \frac{S_2}{M_f L_g}
=
  \frac{S_3^2}{2 (M_f^2 L_g^2 + M_g^4 L_f^2)}
=
  \sqrt{\frac{3D_0}{2L_\Phi^2 \varepsilon^2}}
  ,
$$

the IFO complexity in~\eqref{IFObound2}is
$$
\frac{D_0}{\varepsilon^2}
 + 
\sqrt{D_0}
 \cdot
\left( 1 + \sqrt[4]{\frac{8\varepsilon^2}{3D_0}}  \right) L_\Phi
\cdot
  \frac{\sqrt{216} [ \Phi(x_0) - \Phi^\ast] }{\varepsilon^3}
,
$$
proving~\eqref{IFObound_optimal2}.
\end{proof}

\section{Detailed Proofs of Auxiliary Lemmas}\label{sec:app_aux}
\subsection{Proof of Lemma \ref{lemm:approxg}}\label{sec:proof,lemm:approxg}

\begin{proof}[Proof of Lemma \ref{lemm:approxg}]
We prove the lemma for the case of $t<q$, and for other $t$ it applies directly due to Markov property when epochs (as vectors) are viewed as states of the Markov chain.
\begin{enumerate}[(i)]
\item
We first bound the $\EE\| \bg_t - g(x_t) \|^2$ term and show that for any fixed $t \ge 0$
\beq\label{approxg_bdd}
\EE\| \bg_t - g(x_t) \|^2
 \le
\EE\| \bg_{\lfloor t/q \rfloor q} - g(x_{\lfloor t/q \rfloor q}) \|^2
+
\frac{1}{S_1}
\sum_{s=\lfloor t/q \rfloor q + 1}^t \EE \| g_{j_{t}}\left(x_{s}\right)
-
g_{j_{t}}\left(x_{s-1}\right) \|^2
.
\eeq
First, we take expectation with respect to $j_{t}$ and get $\EE  g_{j_{t}}(x_t) = g(x_t)$ and have
\begin{align*}
\EE\| \bg_t - g(x_t) \|^2
 &=
\EE\| \bg_{t-1}
   +g_{\cS_{1,t}}\left(x_{t}\right)
   -g_{\cS_{1,t}}\left(x_{t-1}\right)
 - g(x_t) \|^2
 \\&=
\EE\| \bg_{t-1} - g(x_{t-1}) + g(x_{t-1})
  +g_{\cS_{1,t}}\left(x_{t}\right)
  -g_{\cS_{1,t}}\left(x_{t-1}\right)
 - g(x_t) \|^2
 \\&=
\EE\| \bg_{t-1} - g(x_{t-1}) \|^2
+
  \EE\| g_{\cS_{1,t}}\left(x_{t}\right)
  -g_{\cS_{1,t}}\left(x_{t-1}\right)
  - (g(x_t) - g(x_{t-1})) \|^2
  \\&=
  \EE\| \bg_{t-1} - g(x_{t-1}) \|^2
\\&\quad  
+\frac{1}{S_1^2} \EE\left(
  \sum_{j\in \cS_{1,t}}  \EE\| g_{j}\left(x_{t}\right)  -g_{j}\left(x_{t-1}\right)
    - (g(x_t) - g(x_{t-1})) \|^2
 \right) 
 \\&\le
\EE\| \bg_{t-1} - g(x_{t-1}) \|^2
  +\frac{1}{S_1} \EE \| g_{j}\left(x_{t}\right)-g_{j}\left(x_{t-1}\right) \|^2
.
\end{align*}
where we used $\EE\|\bX - \EE[ \bX | \cF] \|^2 \le \EE\|\bX\|^2$ for any random vector $\bX$ and any conditional expectation $\EE[\bX | \cF]$.
Apply the above calculations recursively proves \eqref{approxg_bdd}.

\item
We have for the left hand of \eqref{approxg}
\beq\label{eq:11}
\begin{aligned}
\EE\left\|
\nabla \Phi\left(x_t\right) - \left(\partial g\left(x_t\right)\right)^{\top} \nabla f\left(\bg_t\right)
\right\|^2
     &=
\EE\left\|
\left(\partial g\left(x_t\right)\right)^{\top} \nabla f\left(g(x_t)\right)
-
\left(\partial g\left(x_t\right)\right)^{\top} \nabla f\left(\bg_t\right)
\right\|^2
     \\&\le
M_g^2 L_f^2 \EE\|\bg_t - g(x_t)\|^2
     .
\end{aligned}
\eeq
Applying \eqref{approxg_bdd} we obtain
\beq\label{eq:zzjlsdafnlvdsf}
\begin{aligned}
\EE\| \bg_t - g(x_t) \|^2
 &\le
\EE\| \bg_{\lfloor t/q \rfloor q} - g(x_{\lfloor t/q \rfloor q}) \|^2
+
\frac{1}{S_1}
\sum_{s=1}^t
\EE \| g_{j_{t}}\left(x_{s}\right)
-
g_{j_{t}}\left(x_{s-1}\right) \|^2
 \\&\le
\omega_3+
\frac{
M_g^2\cdot \eta^2}{S_1} \sum_{s=1}^t \EE\| \bF_{s-1} \|^2
.
\end{aligned}\eeq
Combining \eqref{eq:zzjlsdafnlvdsf} and \eqref{eq:11} together concludes the proof.
\end{enumerate}
\end{proof}

\subsection{Proof of Lemma \ref{lemm:Gapprox}}\label{sec:proof,lemm:Gapprox}

\begin{proof}[Proof of Lemma \ref{lemm:Gapprox}]
Analogous to Lemma \ref{lemm:approxg} we only consider the case $t<q$.

\begin{enumerate}[(i)]
\item
To begin with, we bound the $\EE \left\|\bG_t - \partial g(x_t)\right\|_F^2$ term (note the Frobenius norm) and conclude
\beq\label{Gapprox_bdd}
\EE \|\bG_t - \partial g(x_t) \|_F^2
 \le
\EE \|\bG_{\lfloor t/q \rfloor q} - \partial g(x_{\lfloor t/q \rfloor q}) \|_F^2
+
\frac{1}{S_2}
\sum_{s=\lfloor t/q \rfloor q+1}^t \EE \| \partial g_{j_{t}}\left(x_{s}\right) - \partial g_{j_{t}}\left(x_{s-1}\right) \|_F^2
.
\eeq
In fact, following the techniques in the proof of \eqref{approxg_bdd} we have
\begin{align*}
\EE \|\bG_t - \partial g(x_t) \|_F^2
 &=
\EE\| \bG_{t-1} +
\partial g_{\cS_{2,t}}\left(x_{t}\right)
-\partial g_{\cS_{2,t}}\left(x_{t-1}\right)
 - \partial g(x_t) \|_F^2
 \\&=
\EE\| \bG_{t-1} - \partial g(x_{t-1}) +\nabla  g(x_{t-1})
+ \partial g_{\cS_{2,t}}\left(x_{t}\right)
- \partial g_{\cS_{2,t}}\left(x_{t-1}\right)
 -\partial g(x_t) \|_F^2
 \\&=
\EE\| \bG_{t-1} -\partial g(x_{t-1}) \|_F^2
\\&\quad+
\EE\| \partial g_{\cS_{2,t}}\left(x_{t}\right)-\partial g_{\cS_{2,t}}\left(x_{t-1}\right)
- (\partial g(x_t) - \partial g(x_{t-1})) \|_F^2
  \\&=
  \EE\| \bG_{t-1} -\partial g(x_{t-1}) \|_F^2
  \\&\quad+ 
\frac{1}{S_2^2}  \EE\left(
  \sum_{i\in\cS_{2,t}} \| \partial g_{j}\left(x_{t}\right)-\partial g_{j}\left(x_{t-1}\right)
  - (\partial g(x_t) - \partial g(x_{t-1})) \|_F^2
  \right)
 \\&\le
\EE\| \bG_{t-1} - \partial g(x_{t-1}) \|_F^2
  + \frac{1}{S_2} \EE \| \partial g_{j}\left(x_{t}\right)-\partial g_{j}\left(x_{t-1}\right) \|_F^2,
\end{align*}
Recursively applying the above gives \eqref{Gapprox_bdd}.

\item
Further, note that for any fixed $t \ge 0$
\beq\label{eq:12}
\EE\left\|
\left(\partial g(x_t)\right)^\top \nabla f(\bg_t)
 - 
\left(\bG_t\right)^\top \nabla f(\bg_t)
\right\|^2
    \le
 M_f^2 \EE \left\|\bG_t - \partial g(x_t)\right\|^2
.
\eeq
Applying \eqref{Gapprox_bdd} we obtain, by smoothness condition \eqref{assmoothness}, that
$$\begin{aligned}
\EE \|\bG_t - \partial g(x_t) \|_F^2
&\le
\EE \|\bG_{\lfloor t/q \rfloor q} - \partial g(x_{\lfloor t/q \rfloor q}) \|_F^2
+ 
\frac{1}{S_2}
\sum_{s=\lfloor t/q \rfloor q+1}^t 
\EE\| \partial g_{j_{t}}\left(x_{s}\right)-\partial g_{j_{t}}\left(x_{s-1}\right) \|_F^2
\\&\le
\omega_2 + \frac{L_g^2\cdot \eta^2}{S_2} \sum_{s=1}^t\EE\|\bF_{s-1}\|_F^2
.
\end{aligned}
$$
Bringing this into~\eqref{eq:12} and note the relation $\|\cdot\| \le \|\bullet\|_F$ for a real matrix we conclude \eqref{Gapprox}.

\end{enumerate}
\end{proof}

\subsection{Proof of Lemma \ref{lemm:Fapprox}}\label{sec:proof,lemm:Fapprox}

\begin{proof}[Proof of Lemma \ref{lemm:Fapprox}]
We prove this for $t<q$, and for other $t$ it follows the same procedure to prove.
This prove is essentially the same reasoning as \eqref{approxg_bdd} and \eqref{Gapprox_bdd}, but is significantly more lengthy due to handling more terms.

First of all, we conclude that for any fixed $t \ge 0$
\beq\label{Fapprox_bdd}
\begin{aligned}
&\hspace{-.2in}
\EE\left\|\bF_t-\left(\bG_t\right)^{\top} \nabla f\left(\bg_t\right)\right\|^2
 \le
\EE\left\|\bF_{\lfloor t/q \rfloor q}-\left(\bG_{\lfloor t/q \rfloor q}\right)^{\top} \nabla f\left(\bg_{\lfloor t/q \rfloor q}\right)\right\|^2
\\& \hspace{1.3in} +
\sum_{s=1}^t
\EE\left\|
\left(\bG_{s}\right)^{\top} \nabla f_{i_{t}}\left(\bg_{s}\right)
-
\left(\bG_{s-1}\right)^{\top} \nabla f_{i_{t}}\left(\bg_{s-1}\right)
\right\|^2
.
\end{aligned}
\eeq
We unfold $\bF_t$ using the update rule to get
\beq\label{eq:lem1:1}
\begin{aligned}
&\mathbb{E}\left\|\bF_{t}-\left(\bG_t\right)^{\top} \nabla f\left(\bg_{t}\right)\right\|^{2}
\\ = &
\mathbb{E}\left\|
\bF_{t-1}
-\left(\bG_{t-1}\right)^{\top} \nabla f_{\cS_{3,t}}\left(\bg_{t-1}\right)
+\left(\bG_{t}\right)^{\top} \nabla f_{\cS_{3, t}}\left(\bg_{t}\right)
-\left(\bG_t\right)^{\top} \nabla f\left(\bg_{t}\right)
\right\|^{2}
.
\end{aligned}
\eeq
Subtracting and adding an auxiliary term $(\bG_t)^\top\nabla f(\bg_t)$, we result in an equivalent expression with the RHS of~\eqref{eq:lem1:1} being
\begin{align*}
&\EE\left\|
\bF_{t-1}-(\bG_{t-1})^{\top}\nabla f(\bg_{t-1})
\right.  \\&+  \left.
  (\bG_{t-1})^\top \nabla f(\bg_{t-1})
-
  \left(\bG_{t-1}\right)^{\top} \nabla f_{\cS_{3,t}}\left(\bg_{t-1}\right)
+
  \left(\bG_{t}\right)^{\top} \nabla f_{\cS_{3,t}}\left(\bg_{t}\right)
-
\left(\bG_t\right)^{\top} \nabla f\left(\bg_{t}\right)
\right\|^2
.
\end{align*}
We note that $\EE\left[
  (\bG_{t-1})^\top \nabla f(\bg_{t-1}) -  \left(\bG_{t-1}\right)^{\top} \nabla f_{\cS_{3,t}}\left(\bg_{t-1}\right) 
 +
 \left(\bG_{t}\right)^{\top} \nabla f_{\cS_{3,t}}\left(\bg_{t}\right) - \left(\bG_t\right)^{\top} \nabla f\left(\bg_{t}\right)
 \right] = 0$.
Taking expectation with respect to $i_t$ before taking total expectation we
result in a recursion:
\beq\label{eq:lem1:2}
\begin{aligned}
& \quad
\mathbb{E}\left\|\bF_{t}-\left(\bG_t\right)^{\top} \nabla f\left(\bg_{t}\right)\right\|^{2}
\\& =
\mathbb{E}\left\|\bF_{t-1}-\left(\bG_{t-1}\right)^{\top} \nabla f\left(\bg_{t-1}\right)\right\|^{2}
\\& \quad
     + \mathbb{E}\left\|\begin{array}{c}(\bG_{t-1})^\top \nabla f(\bg_{t-1})
     -
                          \left(\bG_{t-1}\right)^{\top} \nabla f_{\cS_{3,t}}\left(\bg_{t-1}\right)
     +
                          \left(\bG_{t}\right)^{\top} \nabla f_{\cS_{3,t}}\left(\bg_{t}\right)-\left(\bG_t\right)^{\top} \nabla f\left(\bg_{t}\right)\end{array}\right\|^{2}
                      \\& =
                      \mathbb{E}\left\|\bF_{t-1}-\left(\bG_{t-1}\right)^{\top} \nabla f\left(\bg_{t-1}\right)\right\|^{2}
                      \\&
                      + \frac{1}{S_3^2}\EE\left(
                      \sum_{i\in \cS_{3,t}}\left\|\begin{array}{c}(\bG_{t-1})^\top \nabla f(\bg_{t-1})
                                           -
                                           \left(\bG_{t-1}\right)^{\top} \nabla f_i\left(\bg_{t-1}\right)
                                           +
                                           \left(\bG_{t}\right)^{\top} \nabla f_i\left(\bg_{t}\right)-\left(\bG_t\right)^{\top} \nabla f\left(\bg_{t}\right)\end{array}\right\|^{2}
                                           \right)
\\& \le
 \mathbb{E}\left\|\bF_{t-1}-\left(\bG_{t-1}\right)^{\top} \nabla f\left(\bg_{t-1}\right)\right\|^{2}
 +
  \frac{1}{S_3}  \EE \left\|\left(\bG_{t}\right)^{\top} \nabla f_i\left(\bg_{t}\right)-\left(\bG_{t-1}\right)^{\top} \nabla f_i\left(\bg_{t-1}\right)\right\|^{2}
,
\end{aligned}
\eeq
Applying~\eqref{eq:lem1:2} iteratively from $1$ to $t$ leads to~\eqref{Fapprox_bdd}.

We have
\beq\label{eq:znldkafjdkl}
\begin{aligned}
& \quad
\EE\left\|\bF_t-\left(\bG_t\right)^{\top} \nabla f\left(\bg_t\right)\right\|^2
\\&    \overset{(a)}{\le}
\EE\left\|\bF_{\lfloor t/q \rfloor q}-\left(\bG_{\lfloor t/q \rfloor q}\right)^{\top} \nabla f\left(\bg_{\lfloor t/q \rfloor q}\right)\right\|^2
+
\frac{1}{S_3}
\sum_{s=1}^t
\EE\left\|
\left(\bG_{s}\right)^{\top} \nabla f_{i_{t}}\left(\bg_{s}\right)
-
\left(\bG_{s-1}\right)^{\top} \nabla f_{i_{t}}\left(\bg_{s-1}\right)
\right\|^2
   \\  &  \overset{(b)}{\le}
\omega_1
+
\frac{1}{S_3}
 \sum_{s=1}^t \left(
2M_f^2 \EE\left\|\bG_{s}-\bG_{s-1}\right\|_F^2
+
2 M_g^2 L_f^2 \EE\left\|\bg_{s}-\bg_{s-1}\right\|^2
\right)
  \\  & \overset{(c)}{\le}
\omega_1
+
\frac{2}{S_3} \left(M_g^4L_f^2 + M_f^2 L_g^2\right) \sum_{s=1}^t\EE\left\|x_s-x_{s-1}\right\|^2
 \\&=
\omega_1
+
\frac{2}{S_3} \left(M_g^4L_f^2 + M_f^2 L_g^2\right)\cdot \eta^2 \sum_{s=1}^t \EE\left\|\bF_{s-1}\right\|^2
,
    \end{aligned}
\eeq
where $(a)$ is due to \eqref{Fapprox_bdd}, $(c)$ comes from $\bg$-iteration and $\bG$-iteration in Algorithm \ref{Alg:SARAH-SCGD} as well as Assumptions \ref{Assumption:Smoothness} and \ref{Assumption:Boundedness} on smoothness and boundedness.

The only left is $(b)$, where we utilize \eqref{omegas} and note that for each $s=1, \dots, t$
\beq\label{eqbdd}
\begin{aligned}
&\quad\EE\left\|
 (\bG_s)^\top \nabla f_{i_{t}}(\bg_s) - (\bG_{s-1})^\top \nabla f_{i_{t}}(\bg_{s-1})
\right\|^2
\\&=
\EE\left\|(\bG_s)^\top \nabla f_{i_{t}}(\bg_s) - (\bG_{s-1})^\top \nabla f_{i_{t}}(\bg_s) 
+ (\bG_{s-1})^\top \nabla f_{i_{t}}(\bg_s) - (\bG_{s-1})^\top \nabla f_{i_{t}}(\bg_{s-1}) 
\right\|^2
\\&\le
2\EE \left\| (\bG_{s-1})^\top \nabla f_{i_{t}}(\bg_s) - (\bG_{s-1})^\top \nabla f_{i_{t}}(\bg_{s-1}) \right\|^2
+
2\EE\left\|(\bG_s)^\top \nabla f_{i_{t}}(\bg_s) -(\bG_{s-1})^\top \nabla f_{i_{t}}(\bg_s)\right\|^2 
\\ & \le
2 \left(
M_g^2 L_f^2\EE\left\|\bg_s - \bg_{s-1}\right\|^2
 + 
M_f^2\EE\left\|\bG_s - \bG_{s-1}\right\|_F^2
\right),
\end{aligned}\eeq
which result in $(b)$.
\end{proof}

\end{document}